\documentclass{article}  

\usepackage{ijcai20}

\usepackage{times}
\usepackage{soul}
\usepackage{url}
\usepackage[utf8]{inputenc}
\usepackage[small]{caption}
\usepackage{graphicx}
\usepackage{amsmath}
\usepackage{amsthm}
\usepackage{amssymb}
\usepackage{booktabs}
\usepackage{algorithm}
\usepackage{algorithmic}
\usepackage{color}
\usepackage{subfig}
\usepackage{bbm}
\usepackage{wrapfig}
\DeclareMathOperator*{\softmax}{softmax}

\begin{document}

\title{Efficient Deep Reinforcement Learning via Adaptive Policy Transfer}  

\author{
Tianpei Yang $^{1,2}$\and
Jianye Hao $^{1,2,3}$ \thanks{Corresponding author.} \and
Zhaopeng Meng $^1$ \and 
Zongzhang Zhang $^4$\and
Yujing Hu $^5$ \and \\
Yingfeng Chen $^5$ \and 
Changjie Fan $^5$ \and
Weixun Wang $^1$ \and 
Wulong Liu $^2$ \and
Zhaodong Wang $^6$\and \\
Jiajie Peng $^1$
\affiliations
$^1$College of Intelligence and Computing, Tianjin University\\
 $^2$Noah's Ark Lab, Huawei\\
 $^3$Tianjin Key Lab of Machine Learning\\
 $^4$Nanjing University\\
  $^5$Fuxi AI Lab in Netease\\
  $^6$JD Digits\\
\emails
\{tpyang,jianye.hao,mengzp\}@tju.edu.cn,
zzzhang@nju.edu.cn,
\{huyujing,chenyingfeng1,fanchangjie\}@corp.netease.com,
wxwang@tju.edu.cn,
liuwulong@huawei.com,
zhaodong.wang@jd.com,
jiajiep@gmail.com
}

\maketitle
\begin{abstract}
Transfer Learning (TL) has shown great potential to accelerate Reinforcement Learning (RL) by leveraging prior knowledge from past learned policies of relevant tasks. Existing transfer approaches either explicitly computes the similarity between tasks or select appropriate source policies to provide guided explorations for the target task. However, how to directly optimize the target policy by alternatively utilizing knowledge from appropriate source policies without explicitly measuring the similarity is currently missing. In this paper, we propose a novel Policy Transfer Framework (PTF) to accelerate RL by taking advantage of this idea. Our framework learns when and which source policy is the best to reuse for the target policy and when to terminate it by modeling multi-policy transfer as the option learning problem. PTF can be easily combined with existing deep RL approaches. Experimental results show it significantly accelerates the learning process and surpasses state-of-the-art policy transfer methods in terms of learning efficiency and final performance in both discrete and continuous action spaces.
\end{abstract}

\section{Introduction}\label{sec1}
Recent advance in Deep Reinforcement Learning (DRL) has obtained expressive success of achieving human-level control in complex tasks \cite{mnih2015human,lillicrap2015continuous}. However, DRL is still faced with sample inefficiency problems especially when the state-action space becomes large, which makes it difficult to learn from scratch. TL has shown great potential to accelerate RL \cite{sutton1998reinforcement} via leveraging prior knowledge from past learned policies of relevant tasks \cite{taylor2009transfer,laroche2017transfer,rajendran2017attend}. 
One major direction of transfer in RL focused on measuring the similarity between two tasks either through mapping the state spaces between two tasks \cite{taylor2007transfer,brys2015policy}, or computing the similarity of two Markov Decision Processes (MDPs) \cite{song2016measuring}, and then transferring value functions directly according to their similarities. 

Another direction of policy transfer focuses on selecting a suitable source policy for explorations \cite{fernandez2006probabilistic,li2018optimal}. However, such single-policy transfer cannot be applied to cases when one source policy is only partially useful for learning the target task. Although some transfer approaches utilized multiple source policies during the target task learning, they suffer from either of the following limitations, e.g., Laroche and Barlier \shortcite{laroche2017transfer} assumed that all tasks share the same transition dynamics and differ only in the reward function; Li et al. \shortcite{li2018context} proposed Context-Aware Policy reuSe (CAPS) which required the optimality of source policies since it only learns an intra-option policy over these source policies. Furthermore, it requires manually adding primitive policies to the policy library which limits its generality and cannot be applied to problems of continuous action spaces.


To address the above problems, we propose a novel Policy Transfer Framework (PTF) which combines the above two directions of policy reuse. Instead of using source policies as guided explorations in a target task, we adaptively select a suitable source policy during target task learning and use it as a complementary optimization objective of the target policy. The backbone of PTF can still use existing DRL algorithms to update its policy, and the source policy selection problem is modeled as the option learning problem. In this way, PTF does not require any source policy to be perfect on any subtask and can still learn toward an optimal policy in case none of the source policy is useful. Besides, the option framework allows us to use the termination probability as a performance indicator to determine whether a source policy reuse should be terminated to avoid negative transfer. In summary, the main contributions of our work are: 1) PTF learns when and which source policy is the best to reuse for the target policy and when to terminate it by modelling multi-policy transfer as the option learning problem; 2) we propose an adaptive and heuristic mechanism to ensure the efficient reuse of source policies and avoid negative transfer; and 3) both existing value-based and policy-based DRL approaches can be incorporated and experimental results show PTF significantly boosts the performance of existing DRL approaches, and outperforms state-of-the-art policy transfer methods both in discrete and continuous action spaces. 


\section{Background}\label{sec2}
This paper focuses on standard RL tasks, formally, a task can be specified by an Markov Decision Process (MDP), which can be described as a tuple $<S, A, T, R>$, where $S$ is the set of states; $A$ is the set of actions; $T$ is the state transition function: $S \times A \times S \to [0,1]$ and $R$ is the reward function: $S \times A \times S \to \mathbb{R}$. A policy $\pi$ is a probability distribution over actions conditioned on states: $S \times A \to \left[0,1\right]$. The solution for an MDP is to find an optimal policy $\pi^{*}$ maximizing the total expected return with a discount factor $\gamma$: $U=\sum_{i=t}^{T} \gamma^{i-t} r_i$.

\textbf{Q-Learning, Deep Q-Network (DQN).} \quad Q-learning \cite{watkins1992q} and DQN \cite{mnih2015human} are popular value-based RL methods. Q-learning holds an action-value function for policy $\pi$ as $Q^\pi(s,a) = \mathbb{E}_{\pi}[U|s_t=s, a_t=a]$, and learns the optimal Q-function, which yields an optimal policy \cite{watkins1992q}. DQN learns the optimal Q-function by minimizing the loss:
\begin{equation}\label{q}
L(\theta)=\mathbb{E}_{s,a,r,s'}\left[ \left( r+\gamma \max_{a'}Q'(s',a'|\theta')-Q(s,a|\theta) \right)^2 \right],
\end{equation}
where $Q'$ is the target Q-network parameterized by $\theta'$ and periodically updated from $\theta$.

\textbf{Policy Gradient (PG) Algorithms.} Policy gradient methods are another choice for dealing with RL tasks, which is to directly optimize the policy $\pi$ parameterized by $\theta$. PG methods optimize the objective $J(\theta)=\mathbb{E}_{s\sim P^\pi,a\sim \pi_{\theta}}[U]$ by taking steps in the direction of $\nabla_{\theta}J(\theta)$. Using Q-function, then the gradient of the policy can be written as:
\begin{equation}\label{pg}
\nabla_{\theta}J(\theta)=\mathbb{E}_{s\sim P^\pi,a\sim \pi_{\theta}}[\nabla_{\theta}\log \pi_{\theta}(a|s)Q^{\pi}(s,a)],
\end{equation}
where $P^\pi$ is the state distribution given $\pi$. Several practical PG algorithms differ in how they estimate $Q^{\pi}$. For example, REINFORCE \cite{williams1992simple} simply uses a sample return $U$. Alternatively, one could learn an approximation of the action-value function $Q^{\pi}(s,a)$; $Q^{\pi}(s,a)$ is called the critic and leads to a variety of actor-critic algorithms  \cite{sutton1998reinforcement,mnih2016asynchronous}.

\textbf{The Option Framework.} \quad Sutton et al. \shortcite{sutton1999} firstly formalized the idea of temporally extended actions as an option. An option $o \in \mathcal{O}$ is defined as a triple $\{\mathcal{I}_o, \pi_o, \beta_o\}$ in which $\mathcal{I}_o \in \mathcal{S}$ is an initiation state set, $\pi_o$ is an intra-option policy and $\beta_o: \mathcal{I}_o \to [0,1]$ is a termination function that specifies the probability an option $o$ terminates at state $s \in \mathcal{I}_o$. An MDP endowed with a set of options becomes a Semi-Markov Decision Process (Semi-MDP), which has a corresponding optimal option-value function over options learned using intra-option learning. 
The option framework considers the \textit{call-and-return} option execution model, in which an agent picks option $o$ according to its option-value function $Q(s,o)$, and follows the intra-option policy $\pi_o$ until termination, then selects a next option and repeats the procedure.

\section{Related Work}
Recently, transfer in RL has become an important direction and a wide variety of methods have been studied in the context of RL transfer learning \cite{taylor2009transfer}. Brys et al. \shortcite{brys2015policy} applied a reward shaping approach to policy transfer, benefiting from the theoretical guarantees of reward shaping. However, it may suffer from negative transfer. Song et al. \shortcite{song2016measuring} transferred the action-value functions of the source tasks to the target task according to a task similarity metric to compute the task distance. However, they assumed a well-estimated model which is not always available in practice. Later, Laroche et al. \shortcite{laroche2017transfer} reused the experience instances of a source task to estimate the reward function of the target task. The limitation of this approach resides in the restrictive assumption that all the tasks share the same transition dynamics and differ only in the reward function. 

Policy reuse is a technique to accelerate RL with guidance from previously learned policies, assuming to start with a set of available policies, and to select among them when faced with a new task, which is, in essence, a transfer learning approach \cite{taylor2009transfer}. Fern{\'a}ndez et al. \shortcite{fernandez2006probabilistic} used policy reuse as a probabilistic bias when learning the new, similar tasks. Rajendran et al. \shortcite{rajendran2017attend} proposed the A2T (Attend, Adapt and Transfer) architecture to select and transfer from multiple source tasks by incorporating an attention network which learns the weights of several source policies for combination. Li et al. \shortcite{li2018optimal} proposed the optimal source policy selection through online explorations using multi-armed bandit methods. However, most of the previous works select the source policy according to the performance of source policies on the target task, i.e., the utility, which fails to address the problems where multiple source policies are partially useful for learning the target task and even cause negative transfer. 

\begin{figure*}[t]
\centering
\subfloat[PTF]{
\begin{minipage}[t]{.26\linewidth}  
\includegraphics[width=0.8\linewidth]{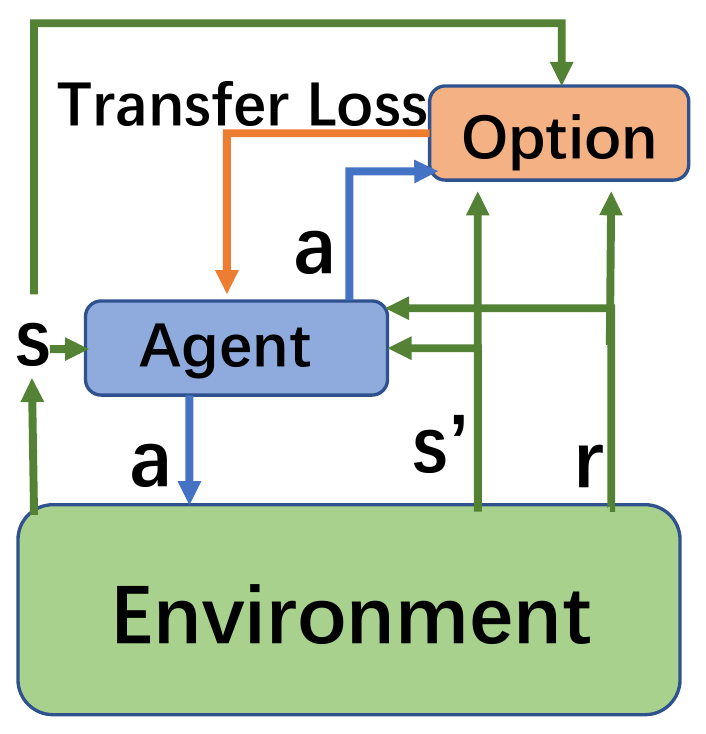}
\end{minipage}
}
\subfloat[Agent Module]{
\begin{minipage}[t]{.28\linewidth}  
\includegraphics[width=0.85\linewidth]{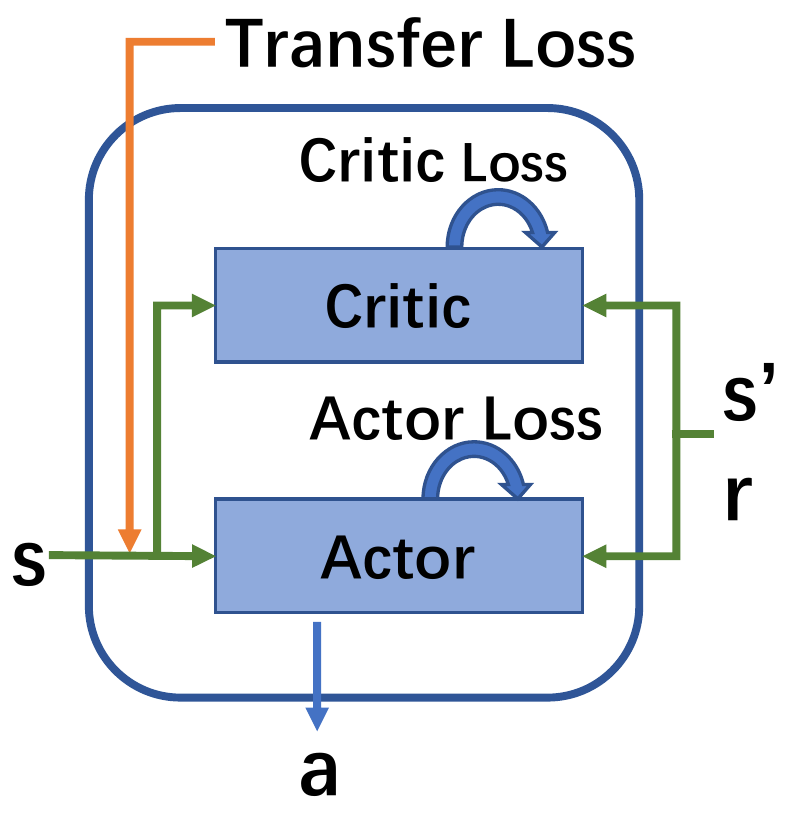}
\end{minipage}
}
\subfloat[Option Module]{
\begin{minipage}[t]{.37\linewidth}  
\includegraphics[width=0.9\linewidth]{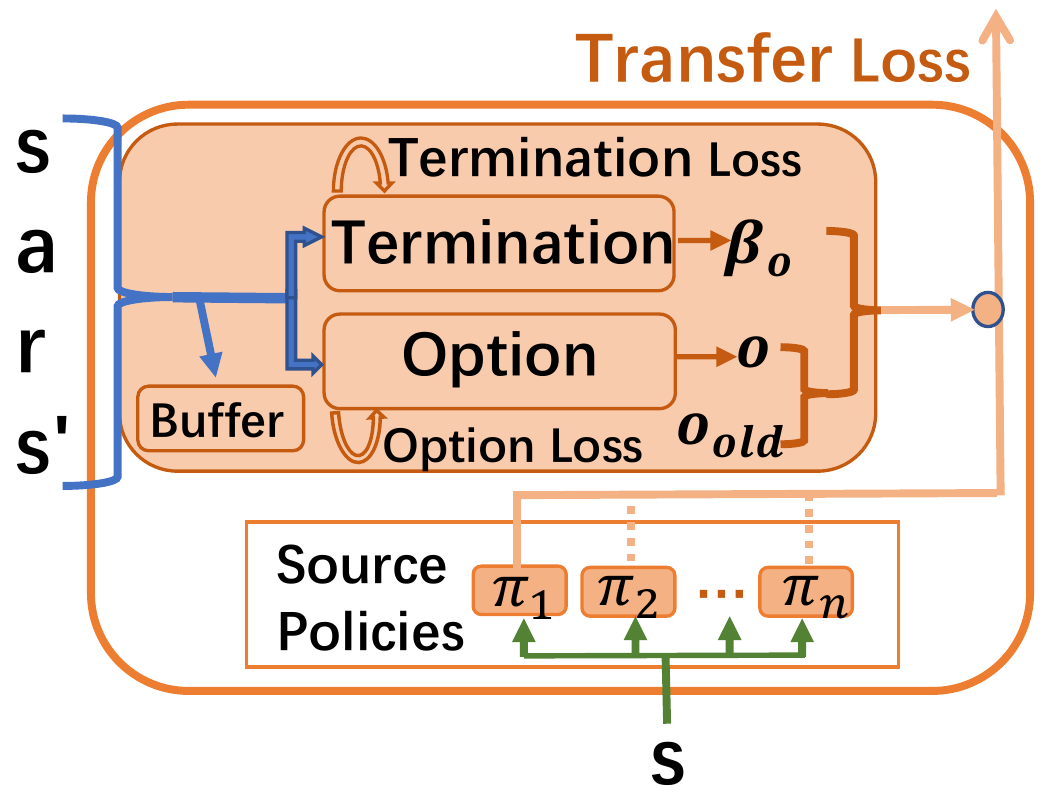}
\end{minipage}
}
\caption{An illustration of the policy transfer framework.}\label{ptf}
\end{figure*}
The option framework was firstly proposed in \cite{sutton1999} as temporal abstractions which is modeled as Semi-MDPs. A number of works focused on option discovery \cite{bacon2017option,abs-1712-00004,HarbBKP18,HarutyunyanDBHM19}. An important example
is the option-critic \cite{bacon2017option} which learns multiple source policies in the form of options from scratch, end-to-end. However, the option-critic tends to collapse to single-action primitives in later training stages. The follow-up work on the option-critic with deliberation cost \cite{HarbBKP18} addresses this option collapse by modifying the termination objective to additionally penalize option termination, but it is highly sensitive to the associated cost parameter. Recently, Harutyunyan et al. \cite{HarutyunyanDBHM19} further modify the termination objective to be completely independent of the task reward and provide theoretical guarantees for the optimality. The objective of all these option discovery works and PTF are orthogonal, that PTF transfers from the source policies to the target task and the rest of works learn multiple source policies from scratch. There are also some imitation learning works \cite{KipfLDZSGKB19,HausmanCSSL17,abs-1711-11289} correlated to option discovery which is not the focus of this work. 

\section{Policy Transfer Framework}\label{sec3}

\subsection{Motivation}\label{sec3.1}
One major direction of previous works focuses on transferring value functions directly according to the similarity between two tasks \cite{brys2015policy,song2016measuring,laroche2017transfer}. However, this way often assumes a well-estimated model for measurement which causes computational complexity and is infeasible in complex scenarios. Another direction of policy transfer methods focuses on selecting appropriate source policies based on the performance of source policies on the target task to provide guided explorations during each episode \cite{fernandez2006probabilistic,li2018optimal,li2018context}. However, most of these works are faced with the challenge of how to select a suitable source policy, since each source policy may only be partially useful for the target task. Furthermore, some of them assume source policies to be optimal and deterministic which restricts the generality. How to directly optimize the target policy by alternatively utilizing knowledge from appropriate source policies without explicitly measuring the similarity is currently missing in previous work.  

According to the above analysis, in this paper, we firstly propose a novel Policy Transfer Framework (PTF) to accelerate RL by taking advantage of this idea and combining the above two directions of policy reuse. Instead of using source policies as guided explorations in a target task, PTF adaptively selects a suitable source policy during target task learning and uses it as a complementary optimization objective of the target policy. In this way, PTF does not require any source policy to be perfect on any subtask and can still learn toward an optimal policy in case none of the source policy is useful. Besides, we propose a novel way of adaptively determining the degree of transferring the knowledge of a source policy to the target one to avoid negative transfer, which can be effectively used in cases when only part of source policies share the same state-action space as the target one. 

\subsection{Framework Overview}\label{sec3.2}

Figure \ref{ptf}(a) illustrates the proposed Policy Transfer Framework (PTF) which contains two main components, one (Figure \ref{ptf}(b)) is the agent module (here is an example of an actor-critic model), which is used to learn the target policy with guidance from the option module. The other (Figure \ref{ptf}(c)) is the option module, which is used to learn when and which source policy is useful for the agent module. Given a set of source policies $\Pi_s=\{\pi_1, \pi_2, \cdots, \pi_{n}\}$ as the intra-option policies, the PTF agent first initializes a set of options $\mathcal{O}=\{o_1, o_2, \cdots, o_{n}\}$ together with the option-value network with random parameters. At each step, it selects an action following its policy, receives a reward and transitions to the next state. Meanwhile, it also selects an option $o_i$ according to the policy over options and the termination probabilities. For the update, the PTF agent introduces a complementary loss, which transfers knowledge from the intra-option policy $\pi_i$ through imitation, weighted by an adaptive adjustment factor $f(\beta_o,t)$. The PTF agent will also update the option-value network and the termination probability of $o_i$ using its own experience simultaneously. The reuse of the policy $\pi_i$ terminates according to the termination probability of $o_i$ and then another option is selected for reuse following the policy over options. In this way, PTF efficiently exploits the useful information from the source policies and avoids negative transfer through the call-and-return option execution model. PTF could be easily integrated with both value-based and policy-based DRL methods. We will describe how it could be combined with A3C \cite{mnih2016asynchronous} as an example in the next section in detail.


\subsection{Policy Transfer Framework (PTF)}\label{sec3.3}
\begin{algorithm}[t]
\caption{PTF-A3C} \label{algo1}
\begin{algorithmic}[1]
\STATE \textbf{Initialize:} option-value network parameters $\theta_o$, termination network parameters $\theta_\beta$, replay buffer $\mathcal{D}$, global parameters $\theta$ and $\theta_{\upsilon}$, thread-specific parameters $\theta'$ and $\theta_{\upsilon}'$, step t $\leftarrow$ 1\\
\FOR{each thread}
\STATE Reset gradients: $d\theta \leftarrow 0, d\theta_{\upsilon} \leftarrow 0$\\
\STATE Assign thread-specific parameters: $\theta' = \theta, \theta_{\upsilon}' = \theta_{\upsilon}$\\
\STATE Start from state $s$, $t_{start} = t$\\
\STATE Select an option $o\leftarrow\epsilon$-greedy$(Q_o(s,o|\theta_o))$\\
\REPEAT
\STATE Perform an action $a \sim \pi(s|\theta')$\\
\STATE Observe reward $r$ and new state $s'$\\
\STATE $t \leftarrow t+1$\\
\STATE Store transition $(s, a, r, s')$ to replay buffer $\mathcal{D}$ \\
\STATE Choose another option if $o$ terminates \\
\UNTIL{$s$ is terminal }\OR{ $t - t_{start} == t_{max}$}
\STATE $R=\begin{cases} 
0& \; \text{if } s \text{ is terminal}\\
V(s, \theta_{\upsilon}') & \; \text{otherwise}
\end{cases}$\\
\FOR{$i \in \{t-1,\cdots, t_{start}\}$}
\STATE $R \leftarrow r_i + \gamma R$\\
\STATE Calculate gradients w.r.t. $\theta_{\upsilon}'$: \\
$d\theta_{\upsilon} \leftarrow d\theta_{\upsilon} + \partial(R-V(s_i|\theta_{\upsilon}'))^2/\partial \theta_{\upsilon}'$\\
\STATE Calculate gradients w.r.t. $\theta'$:\\
$d\theta \leftarrow d\theta+\nabla_{\theta'}\log\pi(a_i|s_i,\theta')(R-V(s_i|\theta_{\upsilon}'))+\rho \nabla_{\theta'}\text{H}(\pi(s_i|\theta'))+f(\beta_o,t)\text{L}_{\text{H}}$\\ 
\STATE Update$(Q_o(s,o|\theta_o))$ (see Algorithm \ref{algo2})\\
\STATE Update $\beta\left( s,o|\theta_{\beta} \right)$ w.r.t. $\theta_\beta$ (Equation \ref{eq3})\\
\ENDFOR
\STATE Asynchronously update $\theta$ using $d\theta$ and $\theta_{\upsilon}$ using $d\theta_{\upsilon}$\\
\ENDFOR
\end{algorithmic}
\end{algorithm}
In this section, we describe PTF applying in A3C \cite{mnih2016asynchronous}: PTF-A3C. The whole learning process of PTF-A3C is shown in Algorithm \ref{algo1}. First, PTF-A3C initializes network parameters for the option-value network, the termination network (which shares the input and hidden layers with the option-value network and holds a different output layer), and A3C networks (Line 1). For each episode, the PTF-A3C agent first selects an option $o$ according to the policy over options (Line 6); then it selects an action following the current policy $\pi(s|\theta')$, receives a reward $r$, transits to the next state $s'$ and stores the transition to the replay buffer $\mathcal{D}$ (Lines 8-11). Another option will be selected if the option $o$ is terminated according to the termination probability of $o$ (Line 12).

For the update, the agent computes the gradient of the temporal difference loss for the critic network (Line 17); and calculates the gradients of the standard actor loss, and also the extra loss of difference between the source policy $\pi_{o}$ inside the option $o$ and the current policy $\pi(\theta')$, which is measured by the cross-entropy loss: $\text{L}_{\text{H}}=\text{H}(\pi_{o}\parallel\pi(\theta'))$. $\text{L}_{\text{H}}$ is used as the supervision signal, weighted by an adaptive adjustment factor $f(\beta_o,t)$. To ensure sufficient explorations, an entropy bonus is also considered \cite{mnih2016asynchronous}, weighted by a constant factor $\rho$ (Line 18). Then it updates the option-value network following Algorithm \ref{algo2} and the termination network accordingly (Lines 19, 20) which is described in detail in the following section.

\subsection{Update the Option Module}\label{sec3.3.1}
The remaining issue is how to update the option-value network which is given in Algorithm \ref{algo2}. Since options are temporal abstractions \cite{sutton1999,bacon2017option}, $U$ is introduced as the option-value function \textsl{upon arrival}. The expected return of executing option $o$ upon entering next state $s'$ is $U(s',o|\theta_o)$, which is correlated to $\beta(s',o|\theta_\beta)$, i.e., the probability that option $o$ terminates in next state $s'$:
\begin{equation}\label{eq1}
\begin{aligned}
U(s',o|\theta_o)\leftarrow &(1-\beta(s',o|\theta_\beta))Q_o'(s',o|\theta_o') + \\ &\beta(s',o|\theta_\beta)\max_{o'\in O}Q_o'(s',o'|\theta_o').
\end{aligned}
\end{equation}
Then, PTF-A3C samples a batch of $N$ transitions from the replay buffer $\mathcal{D}$ and updates the option-value network by minimizing the loss (Line 6 in Algorithm \ref{algo2}). Each sample can be used to update the values of multiple options, as long as the option allows to select the sampled action (for continuous action space, this is achieved by fitting action $a$ in the source policy distribution with a certain confidence interval). Thus the sample efficiency can be significantly improved in an off-policy manner.

\begin{algorithm}[t]
\caption{Update$(Q_o(s,o|\theta_o))$} \label{algo2}
\begin{algorithmic}[1]
\STATE Sample a batch of $N$ transitions $(s, a, r, s')$ from $\mathcal{D}$\\
\FOR{$o\in O$}
\IF{$\pi_{o}$ selects action $a$ at state $s$}
\STATE Update $U(s',o|\theta_o)$ (Equation \ref{eq1}) \\
\STATE Set $y\leftarrow r+\gamma U(s',o|\theta_o)$\\
\STATE Update option by minimizing the loss: \\$L\leftarrow \frac{1}{N}\sum_i(y_i - Q_o(s_i,o|\theta_o))^2$\\
\ENDIF
\ENDFOR
\STATE Copy $\theta_o$ to the target network $Q_o'$ every $\tau$ steps
\end{algorithmic}
\end{algorithm}
PTF-A3C learns option-values in the call-and-return option execution model, where an option $o$ is executed until it terminates at state $s$ based on its termination probability $\beta(s,o|\theta_{\beta})$ and then a next option is selected by a policy over options, which is $\epsilon$-greedy to the option-value $Q_o$. Specifically, with a probability of $1-\epsilon$, the option with the highest option-value is selected (random selection in case of a tie); and PTF-A3C makes random choices with probability $\epsilon$ to explore other options with potentially better performance.

According to the call-and-return option execution model, the termination probability controls when to terminate the current selected option and select another option accordingly. The objective of learning the termination probability is to maximize the expected return $U$, so we update the termination network parameters by computing the gradient of the discounted return objective with respect to the initial condition $(s_1,o_1)$ \cite{bacon2017option}: 
\begin{equation}\label{eq2}
\frac{\partial U(s_1,o_1|\theta_o)}{\partial \theta_\beta} = - \sum_{s',o}\mu(s',o|s_1,o_1)\frac{\partial \beta(s',o|\theta_\beta)}{\partial \theta_{\beta}}A(s',o|\theta_o),
\end{equation}
where $A(s',o|\theta_o)$ is the advantage function which can be approximated as $Q_o(s',o|\theta_o) - \max_{o'\in O}Q_o(s',o'|\theta_o)$, and $\mu(s',o|s_1,o_1)$ is a discounted factor of state-option pairs from the initial condition $(s_1,o_1)$: $\mu(s',o|s_1,o_1) = \sum_{t=0}^{\infty}\gamma^{t}P(s_t=s', o_t=o|s_1,o_1)$. $P(s_t=s', o_t=o|s_1,o_1)$ is the transition probability along the trajectory starting from the initial condition $(s_1,o_1)$ to $(s',o)$ in $t$ steps. Since $\mu(s',o|s_1,o_1)$ is estimated from samples along the on-policy stationary distribution, we neglect it for data efficiency \cite{Thomas14,li2018context}. Then $\beta\left( s,o|\theta_{\beta} \right)$ is updated w.r.t. $\theta_\beta$ as follows \cite{bacon2017option,li2018context}:
\begin{equation}\label{eq3}
\theta_\beta \leftarrow \theta_\beta - \alpha_{\beta}\frac{\partial \beta(s',o|\theta_\beta)}{\partial \theta_{\beta}}\left(A(s',o|\theta_o) + \xi\right),
\end{equation}
where $\alpha_{\beta}$ is the learning rate, $\xi$ is a regularization term. The advantage term is $0$ if the option is the one with the maximized option value, and negative otherwise. In this way, all termination probabilities would increase if the option value is not the maximized one. However, the estimation of the option-value function is not accurate initially. If we multiply the advantage to the gradient, the termination probability of an option with the maximize true option value would also increase, which would lead to a sub-optimal policy over options. The purpose of $\xi$ is to ensure sufficient exploration that the best one could be selected.
\subsection{Transfer from Selected Source Policy}
Next, we describe how to transfer knowledge from the selected source policy. The way to transfer is motivated from policy distillation \cite{rusu2016policy} which exploits multiple teacher policies to train a student policy. Namely, a teacher policy $\pi_t$ is used to generate trajectories $x$, each containing a sequence of states $(x_t)_{t \geq 0}$. The goal is to match student's policy $\pi_s(\theta)$, parameterized by $\theta$, to $\pi_t$. The corresponding loss function term for each sequence at each time step $t$ is: $\text{H}(\pi_t(a|x_t)\parallel\pi_s(a|x_t,\theta))$, where $\text{H}(\cdot \parallel \cdot)$ is the cross-entropy loss. For value-based algorithms, e.g., DQN, we can measure the difference of two Q-value distributions using the Kullback-Leibler divergence (KL) with temperature $\tau$: 
\begin{equation}
\text{KL}=\sum_{i=1}^{|\mathcal{D}|}\softmax\left( \frac{\mathbf{q}_{t}(s_i)}{\tau}\right) \ln\frac{\softmax\left( \frac{\mathbf{q}_{t}(s_i)}{\tau}\right)}{\softmax\left(\mathbf{q}_{s}(s_i)\right)}.
\end{equation} 
Kickstarting \cite{schmitt2018kickstarting} trains a student policy that surpasses the teacher policy on the same task set by adding the cross-entropy loss between the teacher and student policies to the RL loss. However, it does not consider learning a new task that is different from the teacher's task set. Furthermore, the way using Population Based Training (PBT) \cite{pbt} to adjust the weighting factor of the cross-entropy loss increases the computational complexity, lack of adaptive adjustment.

To this end, we propose an adaptive and heuristic way to adjust the weighting factor $f(\beta_o)$ of the cross-entropy loss. The option module contains a termination network that reflects the performance of options on the target task. If the performance of the current option is not the best among all options, the termination probability of this option grows, which indicates we should assign a higher probability to terminate the current option. Therefore, the termination probability of a source policy can be used as a performance indicator of adjusting its exploitation degree. Specifically, the probability of exploiting the current source policy $\pi_o$ should be decreased as the performance of the option $o$ decreases. And the weighting factor $f(\beta_o,t)$ which implies the probability of exploiting the current source policy $\pi_o$ should be inversely proportional to the termination probability. Specifically, we propose adaptively adjust $f(\beta_o,t)$ as follows:
\begin{equation}\label{eqt}
f(\beta_o,t) = f(t)(1-\beta(s_t,o|\theta_\beta)),
\end{equation}
where $f(t)$ is a discount function. When the value of the termination function of option $o$ increases, it means that the performance of the option $o$ is not the best one among all options based on the current experience. Thus we decrease the weighting factor $f(\beta_o,t)$ of the cross-entropy loss $\text{H}(\pi_{o}\parallel\pi(\theta))$ and vice versa. $f(t)$ controls the slow decrease in exploiting the transferred knowledge from source policies which means at the beginning of learning, we exploit source knowledge mostly. As learning continues, past knowledge becomes less useful and we focus more on the current self-learned policy. 
In this way, PTF efficiently exploits useful information and avoids negative transfer from source policies.

\section{Experimental Results}\label{sec4}
\begin{figure}[t]
\centering
\subfloat[Grid world $W$]{
\includegraphics[width=0.47\linewidth]{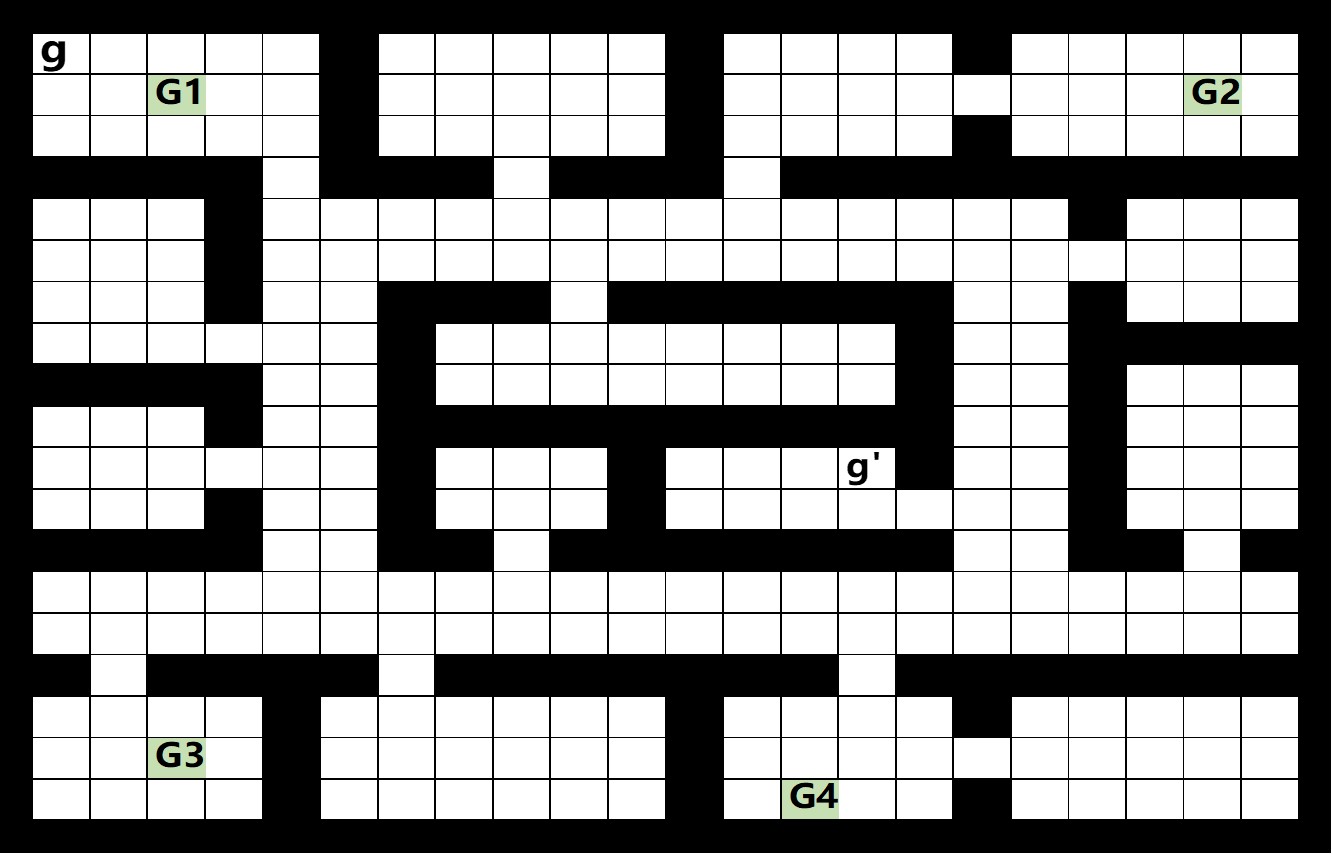}
}
\subfloat[Grid world $W'$]
{
\includegraphics[width=0.47\linewidth]{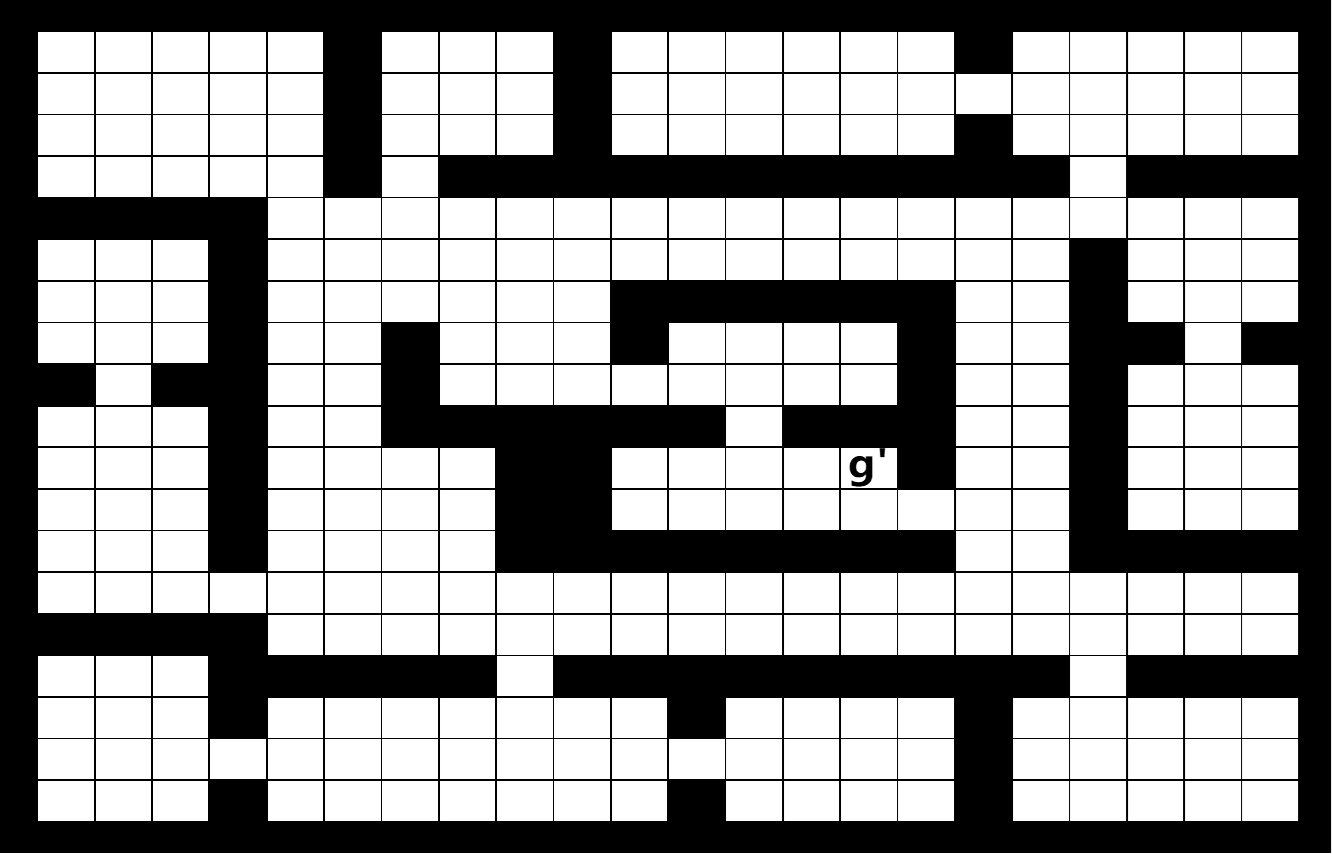}
}
\caption{Two grid worlds. (a) $W$ contains two target tasks $g$, $g'$, four source tasks; (b) $W'$ contains the same target task $g'$.}\label{grid}
\end{figure}
\begin{figure*}[t]
\centering
   \subfloat[A3C vs PTF-A3C]{  
	\includegraphics[width=.29\linewidth]{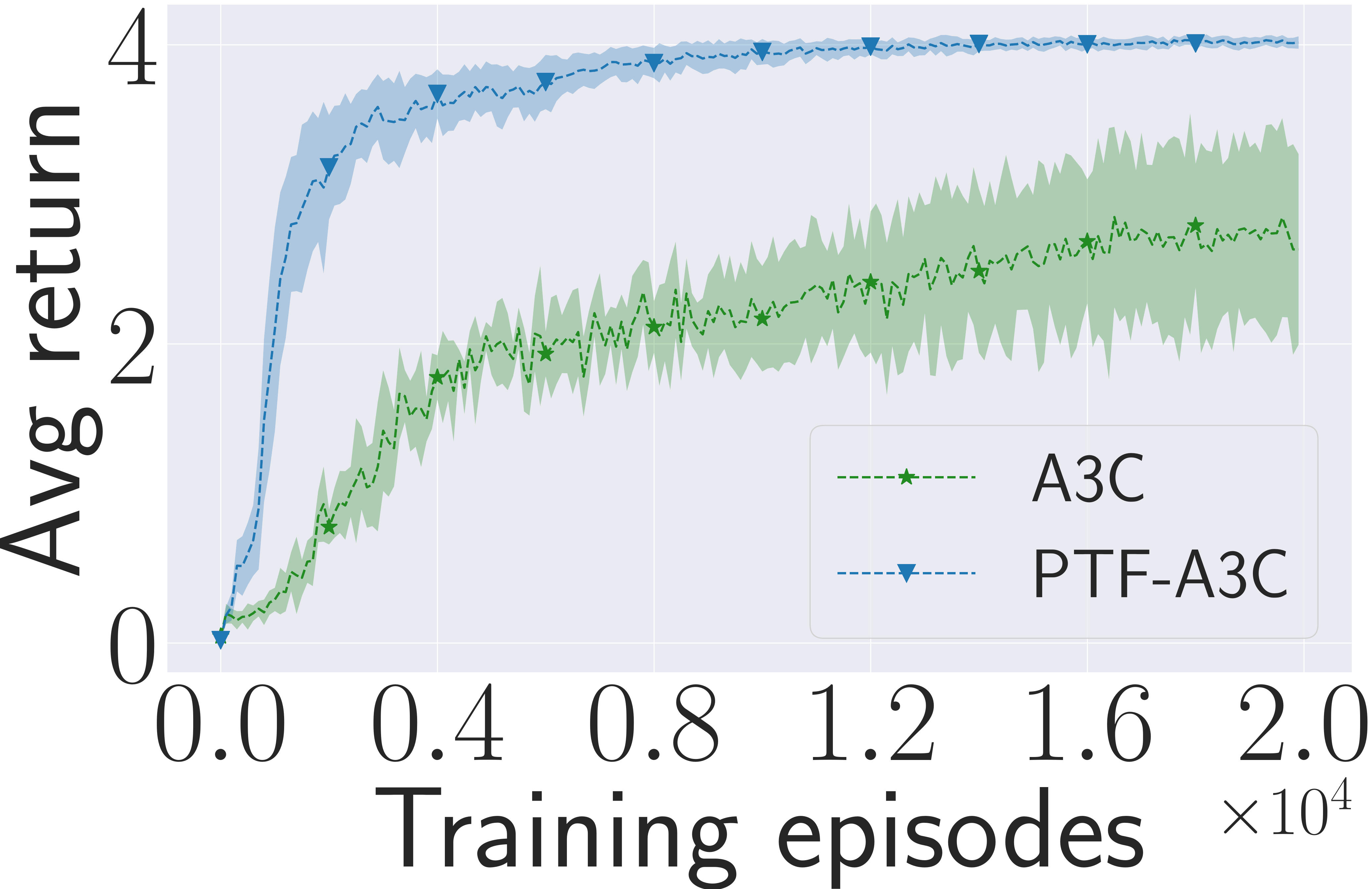}
	}
 \subfloat[PPO vs PTF-PPO]{ 
	\includegraphics[width=.29\linewidth]{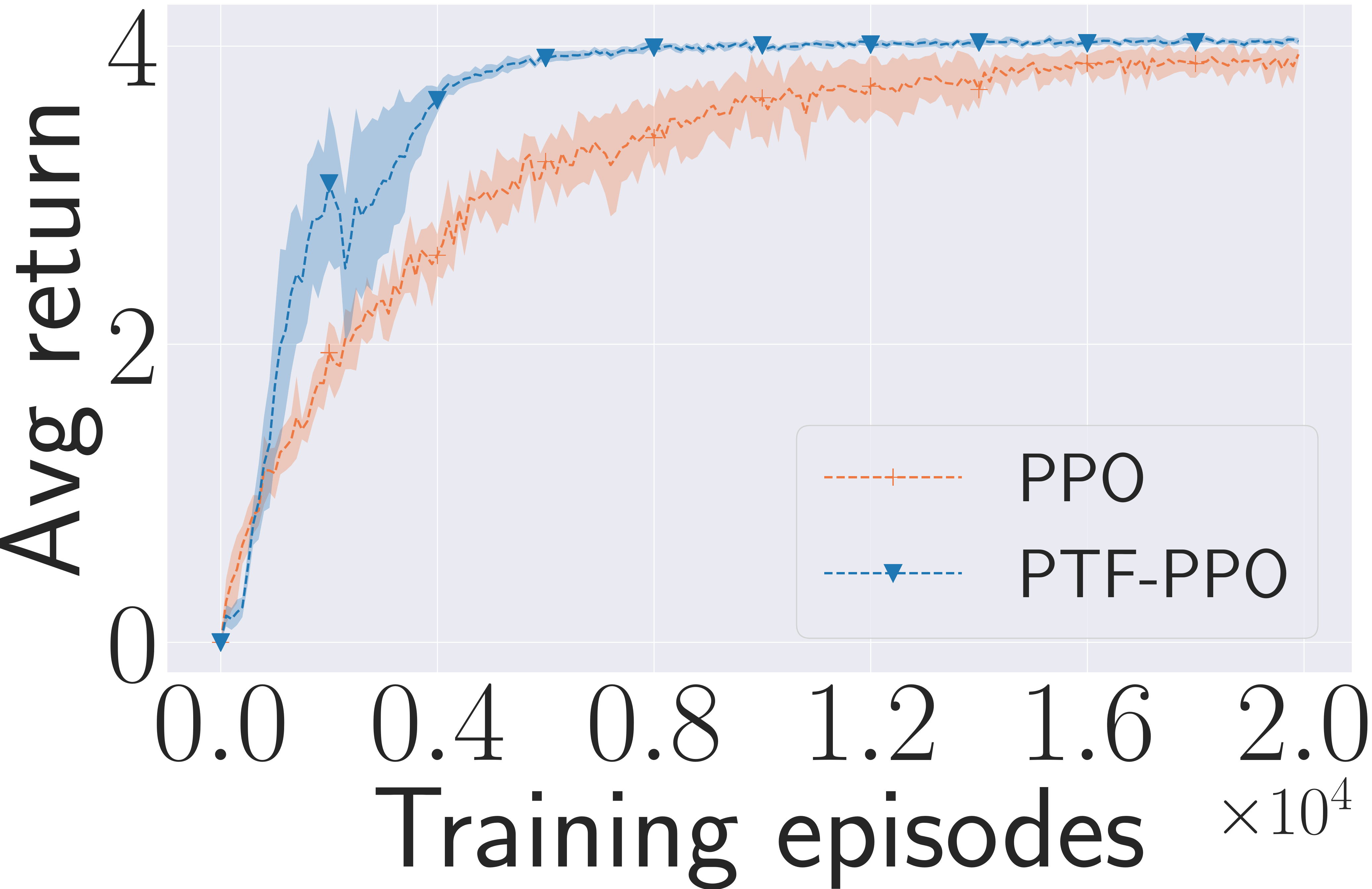}
  	} 
 \subfloat[Deep-CAPS vs PTF-A3C]{
  	\includegraphics[width=.29\linewidth]{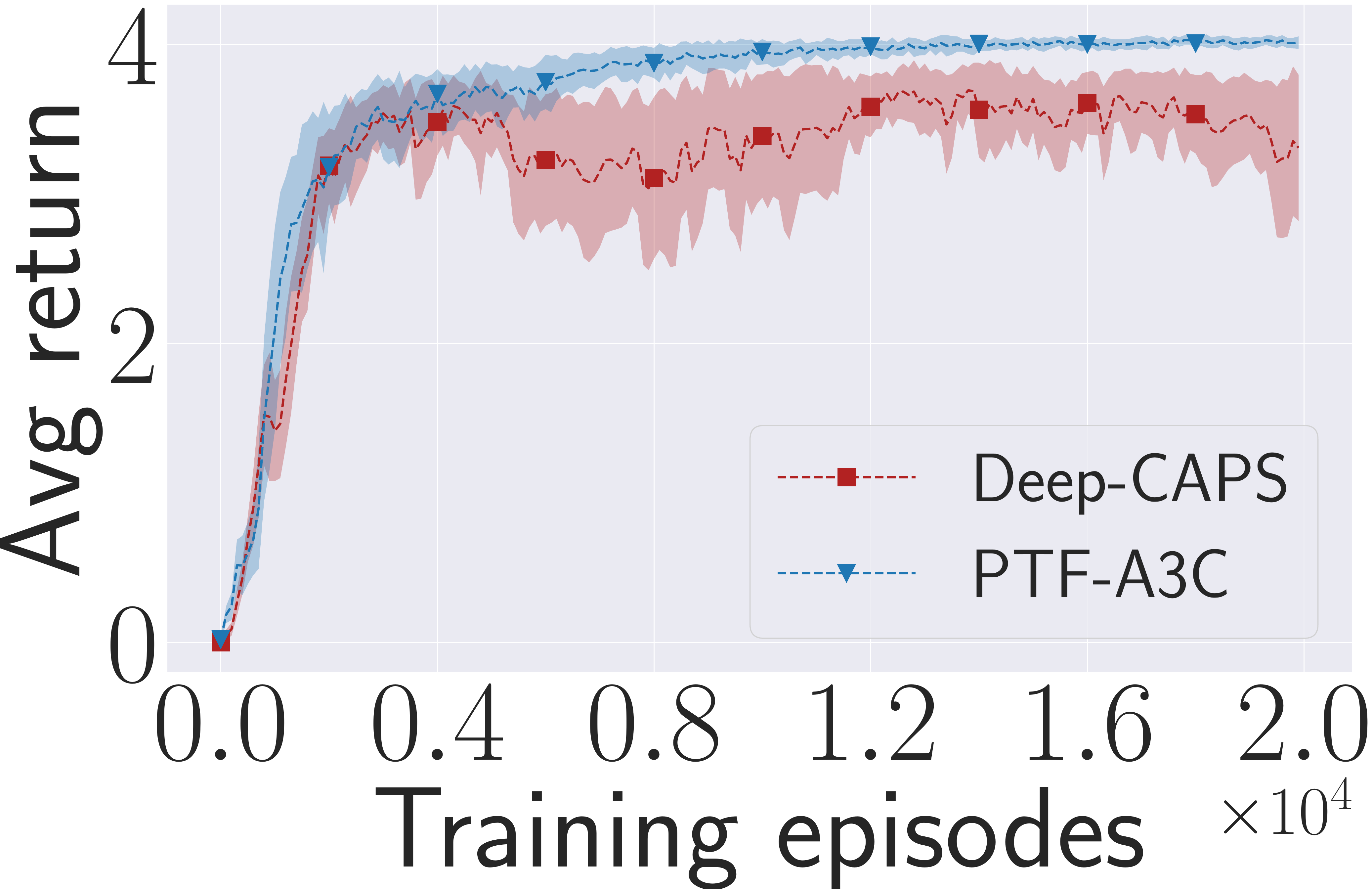}	
  	}
\caption{Average discounted rewards of various methods when learning task $g$ on grid world $W$.} \label{fig-grid2}
\end{figure*}

\begin{figure*}[t]
\centering
   \subfloat[A3C vs PTF-A3C]{
  \includegraphics[width=.29\linewidth]{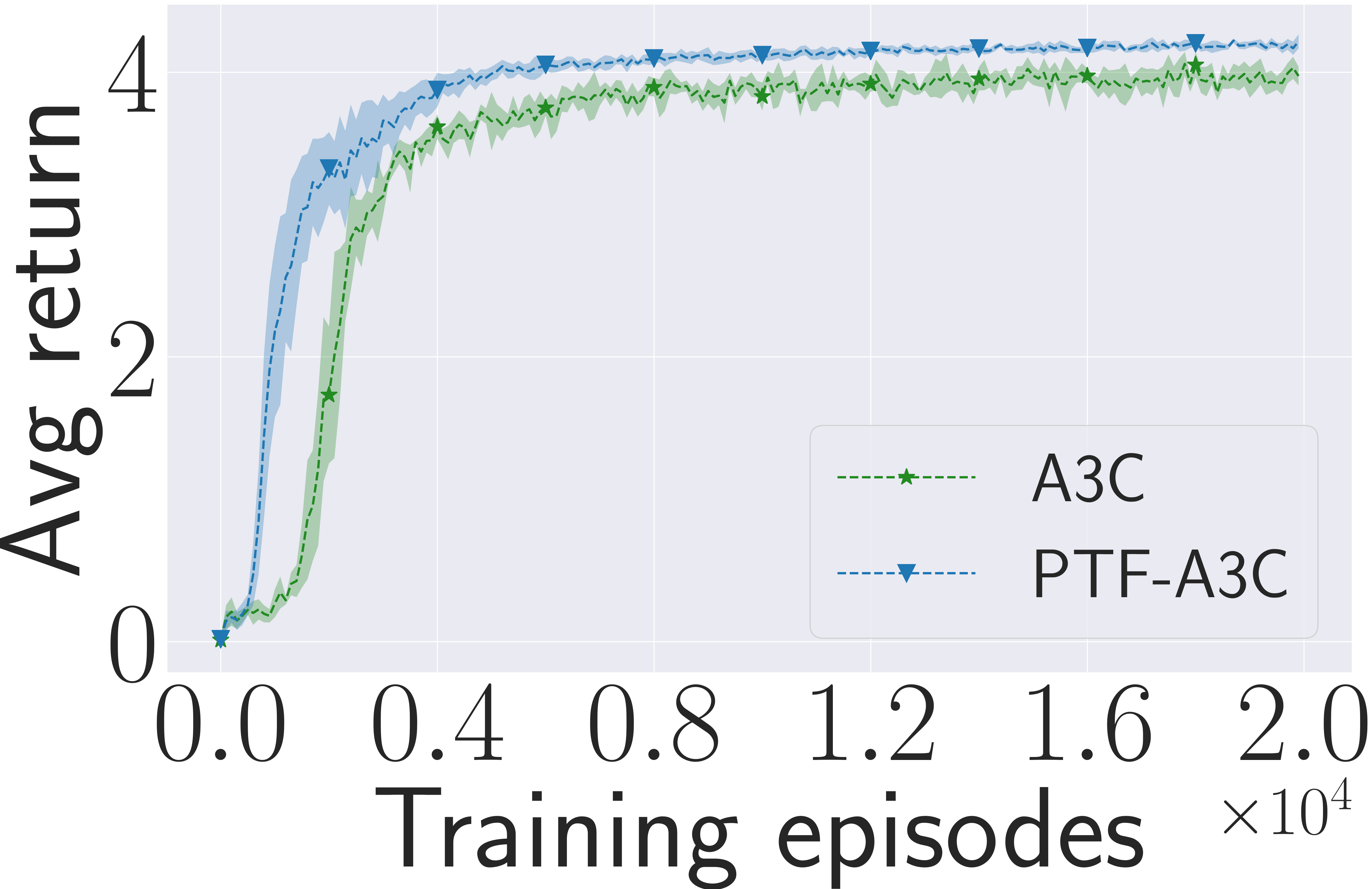}
  }
 \subfloat[PPO vs PTF-PPO]{    
	\includegraphics[width=.29\linewidth]{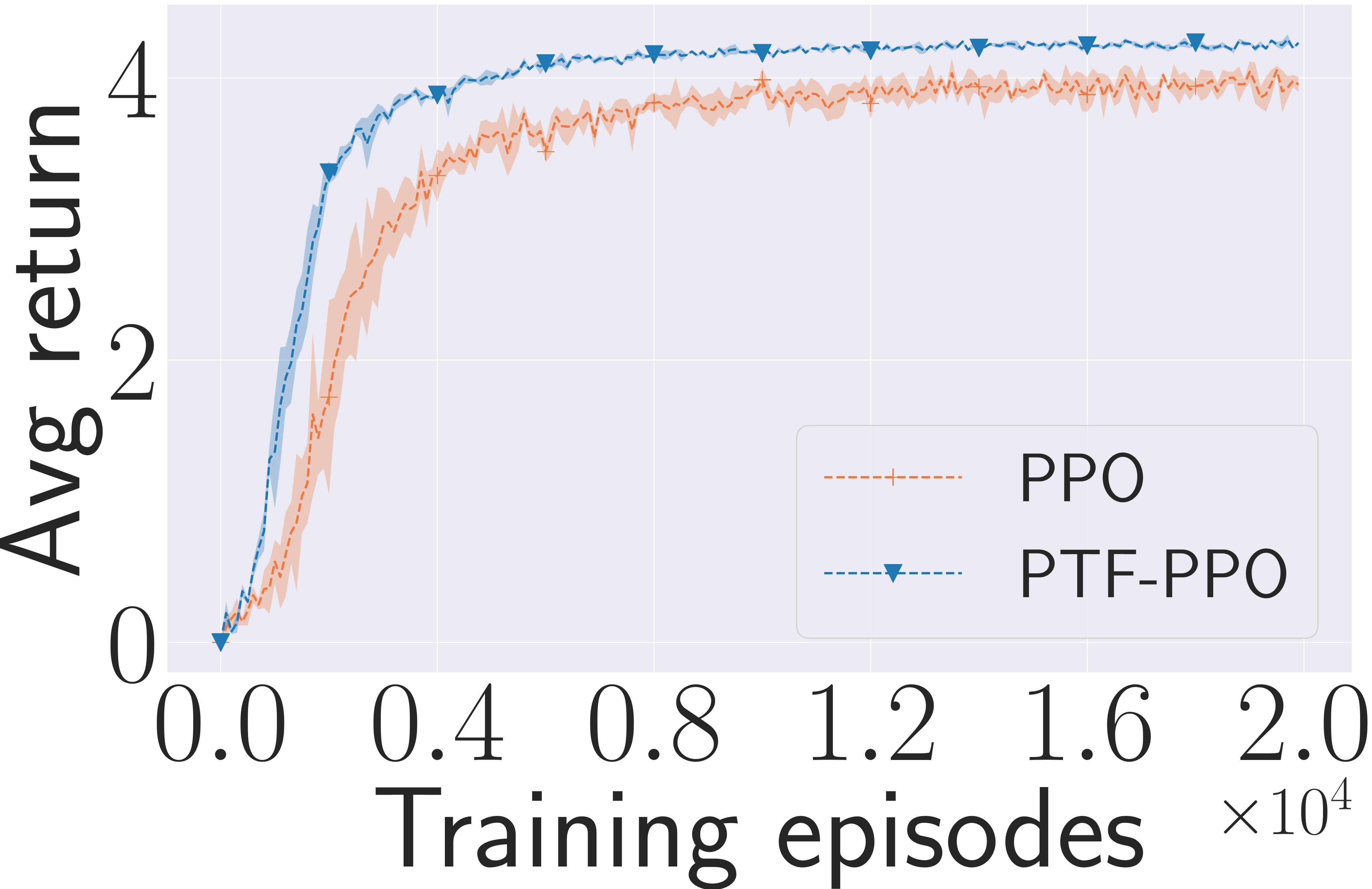}	
  } 
   \subfloat[Deep-CAPS vs PTF-A3C]{    
	\includegraphics[width=.29\linewidth]{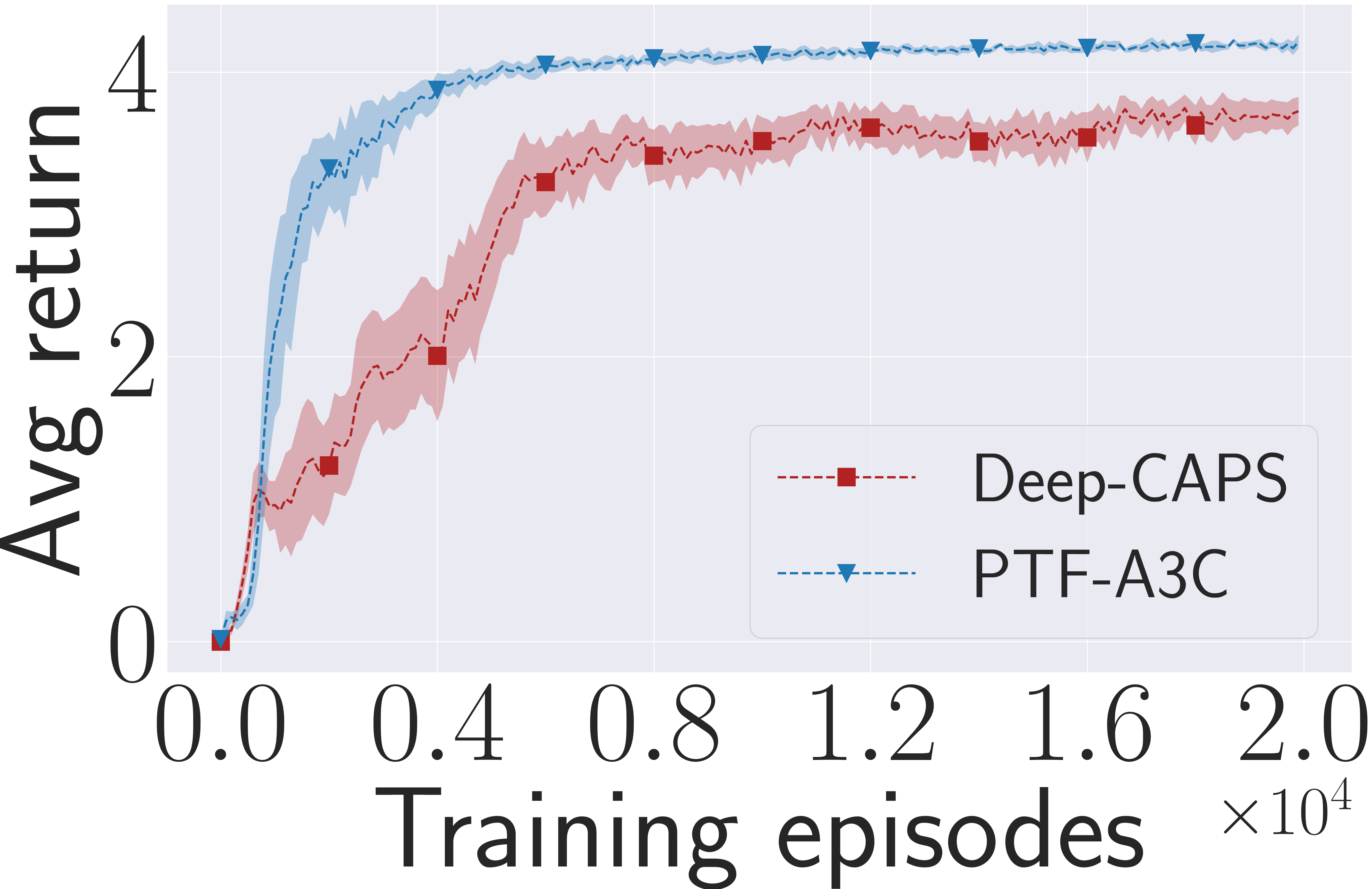}
  }
\caption{Average discounted rewards of various methods when learning task $g'$ on grid world $W$.} \label{fig-grid}
\end{figure*}
In this section, we evaluate PTF on three test domains, grid world \cite{fernandez2006probabilistic}, pinball \cite{konidaris2009skill} and reacher \cite{dmcontrol} compared with several DRL methods learning from scratch (A3C \cite{mnih2016asynchronous} and PPO \cite{ppo}); and the state-of-the-art policy transfer method CAPS \cite{li2018context}, implemented as a deep version (Deep-CAPS). Results are averaged over $20$ random seeds \footnote{The source code is put on \url{https://github.com/PTF-transfer/Code_PTF}}.


\subsection{Grid world}\label{sec4.1.2}

Figure \ref{grid}(a) shows a $24 \times 21$ grid world $W$, with an agent starting from any of the grids, and choosing one of four actions: up, down, left and right. Each action makes the agent move to the corresponding direction with one step size. 
$G1,G2,G3\text{ and }G4$ denote goals of source tasks, $g$ and $g'$ represent goals of target tasks. As noted, $g$ is similar to one of the source tasks $G1$ since their goals are within a close distance; while $g'$ is different from each source task due to the far distance among their goals. 
The game ends when the agent approaches the grid of a target task or the time exceeds a fixed period. The agent receives a reward of $+5$ after approaching the goal grid. The source policies are trained using A3C learning from scratch. We also manually design $4$ primitive policies for deep-CAPS following its previous settings (i.e., each primitive policy selects the same action for all states), which is unnecessary for our PTF framework.

We first investigate the performance of PTF when the target task $g$ is similar to one of the source tasks, $G1$ (i.e., the distance between their goal grids is very close). Figure \ref{fig-grid2} presents the average discounted rewards of various methods when learning task $g$ on grid world. We can see from Figure \ref{fig-grid2}(a) that PTF-A3C significantly accelerates the learning process and outperforms A3C. Similar results can be found in Figure \ref{fig-grid2}(b). The reason is that PTF quickly identifies the optimal source policy and exploits useful information from source policies, which efficiently accelerates the learning process than learning from scratch. Figure \ref{fig-grid2}(c) shows the performance gap between PTF-A3C and deep-CAPS. This is because the policy reuse module and the target task learning module in PTF are loosely decoupled, apart from reusing knowledge from source policies, PTF is also able to utilize its own experience from the environment. However, in deep-CAPS, these two parts are highly decoupled, which means its explorations and exploitations are fully dependent on the source policies inside the options. Thus, deep-CAPS needs higher requirements on source policies than our PTF, and finally achieves lower performance than PTF-A3C.

\begin{figure*}[t]
\centering
\subfloat[A3C vs PTF-A3C]{  
	\includegraphics[width=.29\linewidth]{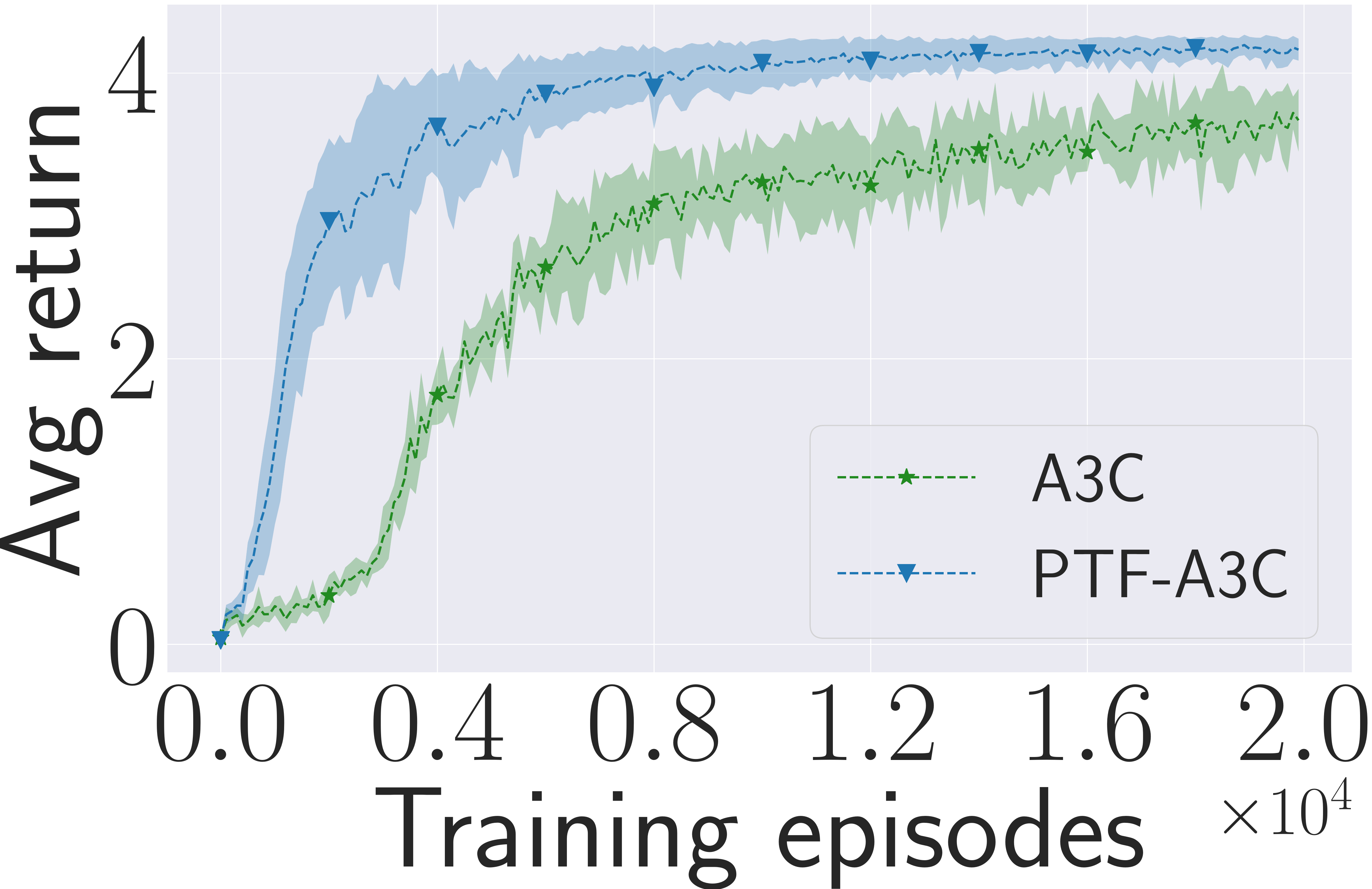}
  }
   \subfloat[PPO vs PTF-PPO]{ 
	\includegraphics[width=.29\linewidth]{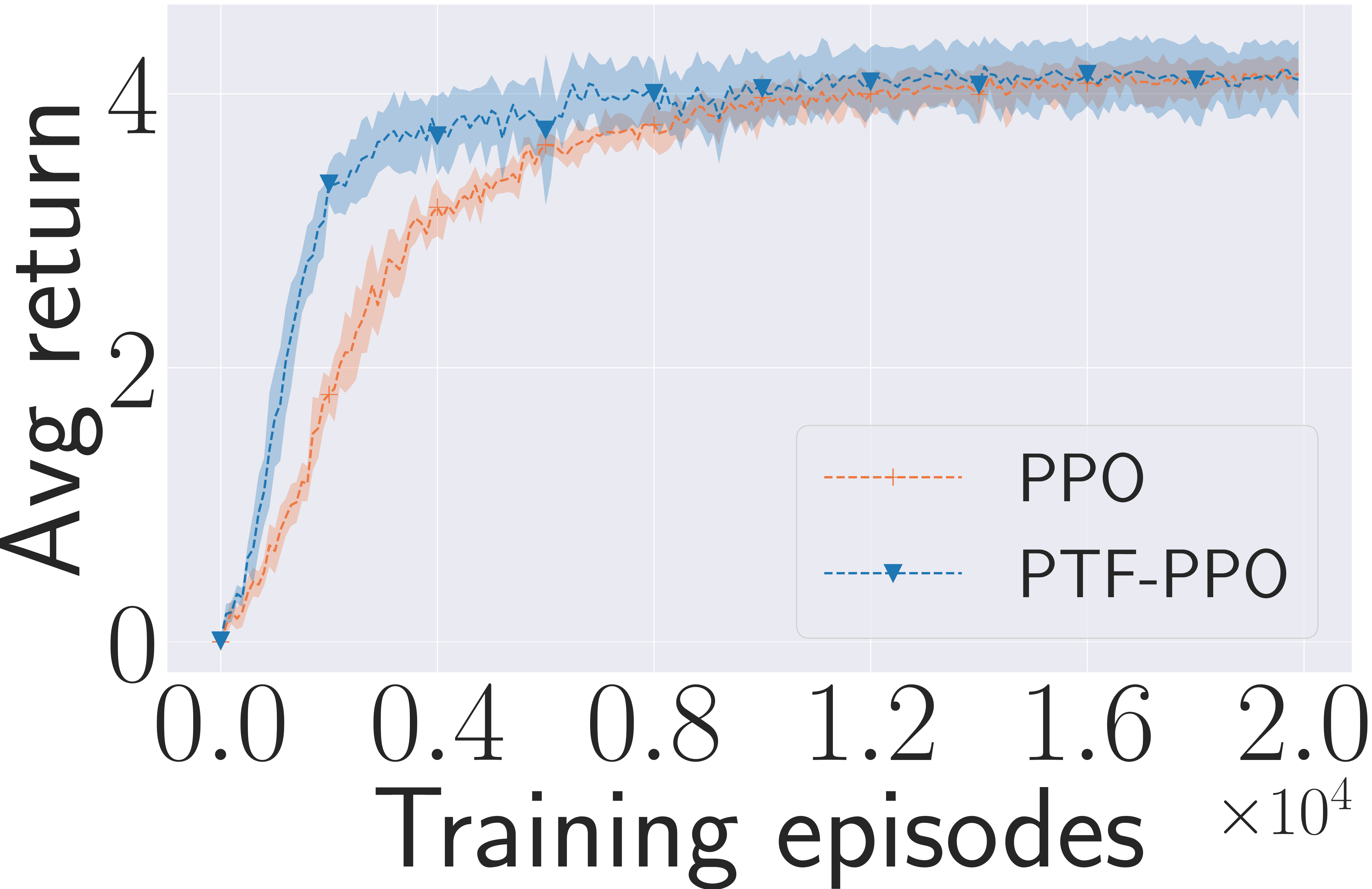}
  } 
   \subfloat[Deep-CAPS vs PTF-A3C]{  
	\includegraphics[width=.29\linewidth]{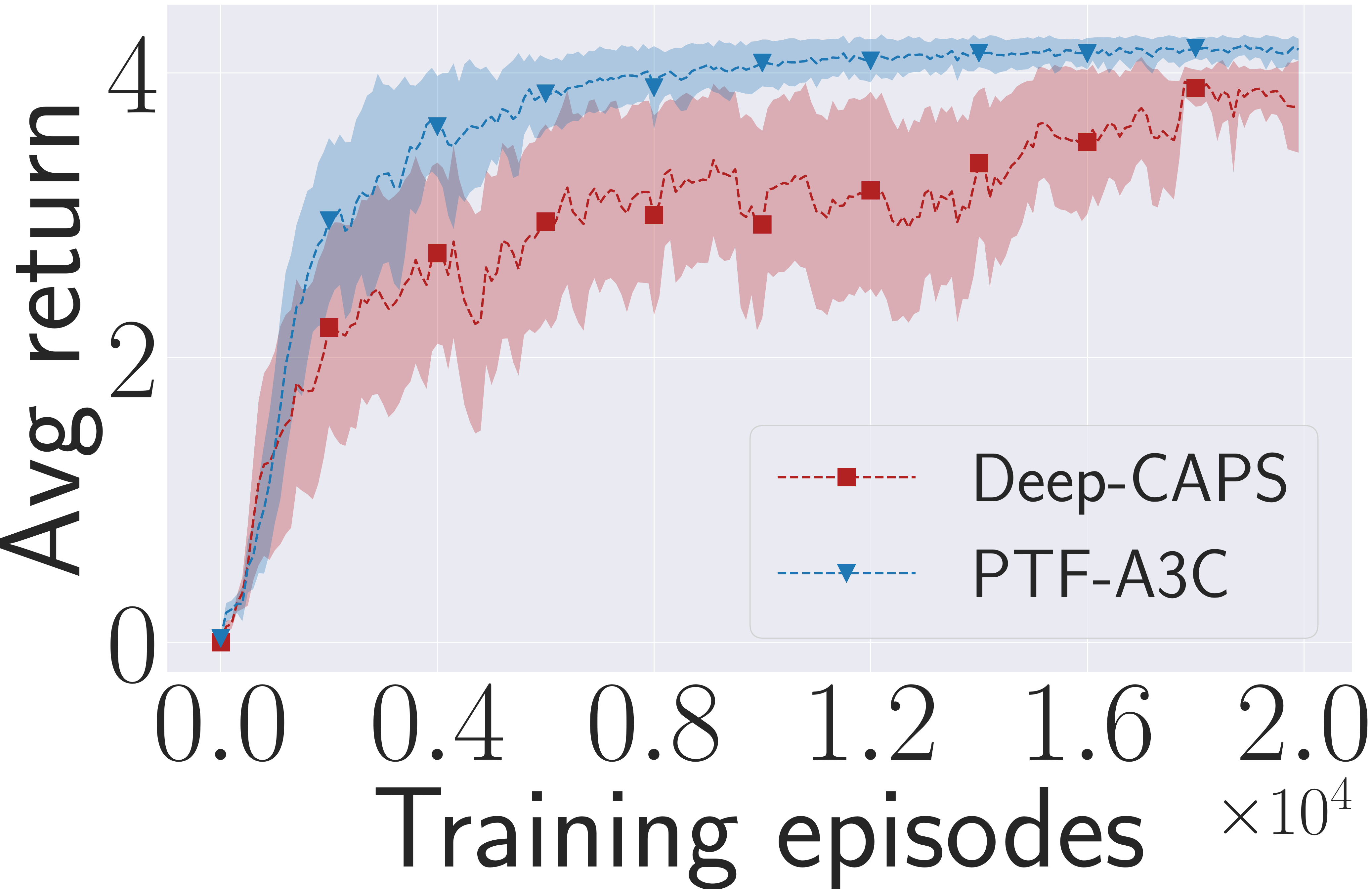}
  }
\caption{Average discounted rewards of various methods when learning task $g'$ on grid world $W'$.} \label{fig-grid3}
\end{figure*}
Next, we investigate the performance of PTF when all source tasks are not quite similar to the target task (i.e., the distance between their goal grids is very far). Figure \ref{fig-grid} presents average discounted rewards of various methods when learning task $g'$. We can see from Figure \ref{fig-grid}(a), (b) that both PTF-A3C and PTF-PPO significantly accelerate the learning process and outperform A3C and PPO. The reason is that PTF identifies which source policy is optimal to exploit and when to terminate it, which efficiently accelerates the learning process than learning from scratch. The lower performance of deep-CAPS than PTF-A3C (Figure \ref{fig-grid}(c)) is due to the similar reasons as described before, that its explorations and exploitations are fully dependent on source policies, thus needs higher requirements on source policies than PTF, and finally achieves lower performance than PTF-A3C.

To verify that PTF works as well in situations where transitions between source and target tasks are different, we conduct experiments on learning on a grid world $W'$ (Figure \ref{grid}(b)), whose map is much different from the map for learning source tasks. Figure \ref{fig-grid3} shows that PTF still outperforms other methods even if only some parts of source policies can be exploited. PTF identifies and exploits useful parts automatically.

\begin{figure}[t]
\centering
\includegraphics[width=0.6\linewidth]{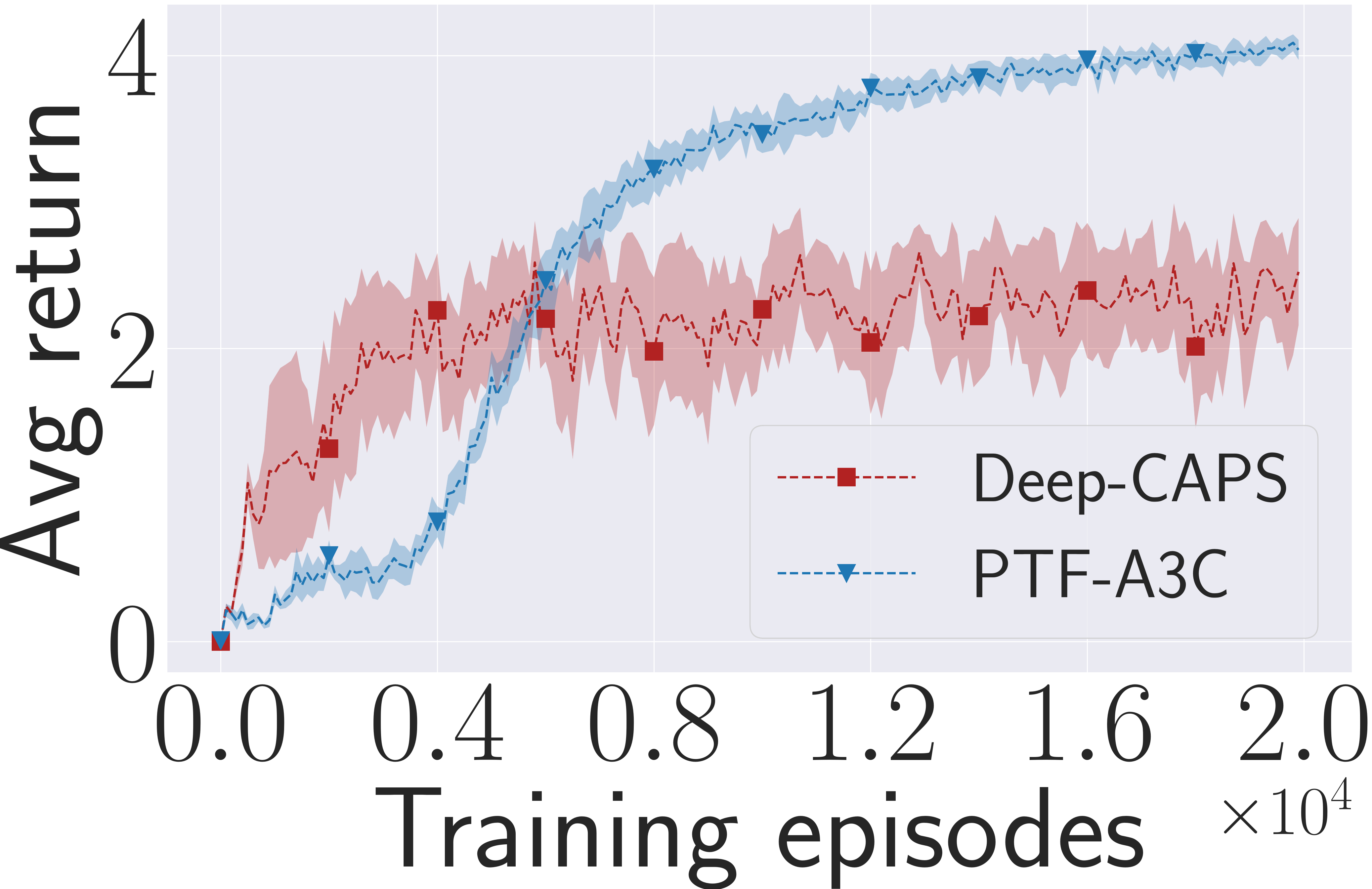}
\caption{The performance of PTF-A3C and deep-CAPS on grid world $W$ with imperfect source policies.} \label{fig-negative}
\end{figure}
We further investigate whether PTF can efficiently avoid negative transfer. Figure \ref{fig-negative} shows the average discounted rewards of PTF-A3C and deep-CAPS when source policies are not optimal towards source tasks. As we described before, deep-CAPS is fully dependent on source policies for explorations and exploitations on the target task, when source policies are not optimal towards source tasks, which means they are not deterministic at all states. Thus, deep-CAPS cannot avoid the negative and stochastic impact of source policies, which confuses the learning of the option-value network and finally obtains lower performance than PTF-A3C.

\subsection{Pinball}\label{sec4.1.3}

\begin{figure}[t]
\centering
\subfloat[Pinball]{
\includegraphics[width=0.45\linewidth , height=1.3in]{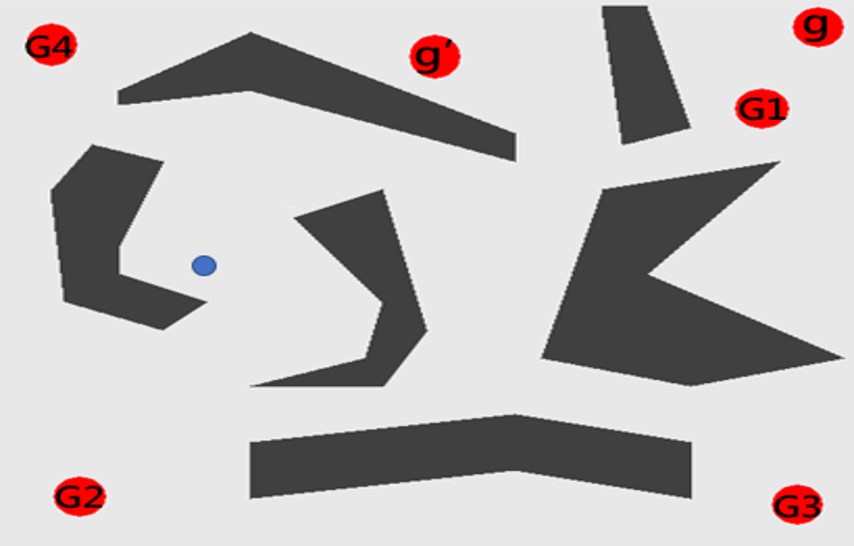}
}
\subfloat[Reacher]{
\includegraphics[width=0.45\linewidth , height=1.3in]{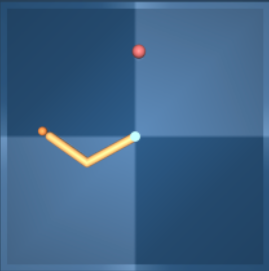}
}
\caption{Two evaluation environments with continuous control.}\label{pinball}
\end{figure}

In the pinball domain (Figure \ref{pinball}(a)), a ball must be guided through a maze of arbitrarily shaped polygons to a designated target location. The state space is continuous over the position and velocity of the ball in the $x-y$ plane. The action space is continuous in the range of $[-1,1]$, which controls the increment of the velocity in the vertical or horizontal direction. Collisions with obstacles are elastic and can be used to the advantage of the agent. A drag coefficient of $0.995$ effectively stops ball movements after a finite number of steps when the null action is chosen repeatedly. Each thrust action incurs a penalty of $-5$ while taking no action costs $-1$. The episode terminates with a $+10000$ reward when the agent reaches the target. We interrupted any episode taking more than $500$ steps and set the discount factor to $0.99$. These rewards are all normalized to ensure more stable training. The source policies are trained using A3C learning from scratch. We also design $5$ primitive policies for deep-CAPS, an increment $+1$ of the velocity in the vertical or horizontal direction; a decrement $-1$ of the velocity in the vertical or horizontal direction and the null action, which is unnecessary for our PTF framework. 
\begin{figure*}
\centering
   \subfloat[A3C vs PTF-A3C]{
   \includegraphics[width=.29\linewidth]{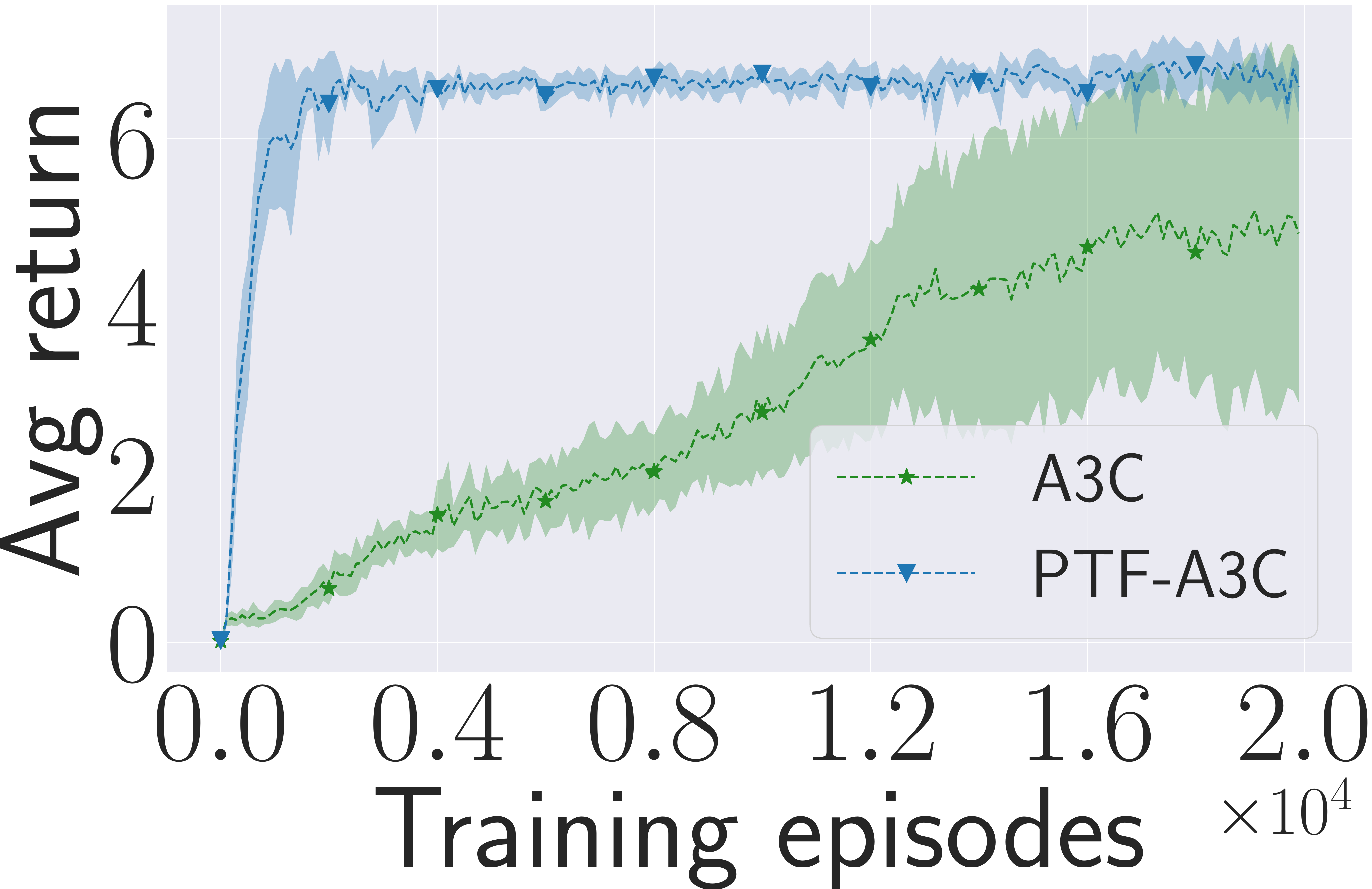}
  }
   \subfloat[PPO vs PTF-PPO]{
  \includegraphics[width=.29\linewidth]{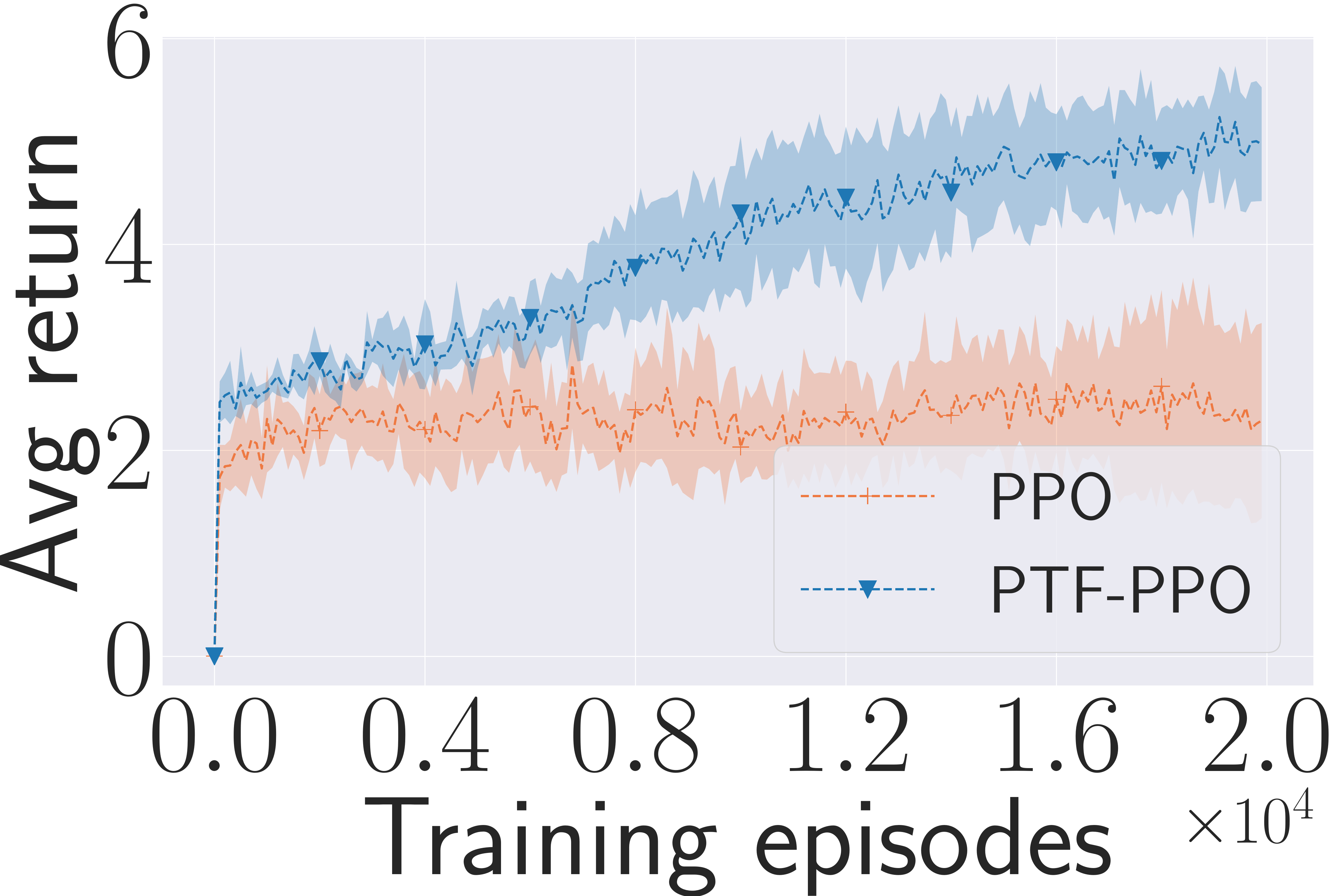}	
  } 
   \subfloat[Deep-CAPS vs PTF-A3C]{
  \includegraphics[width=.29\linewidth]{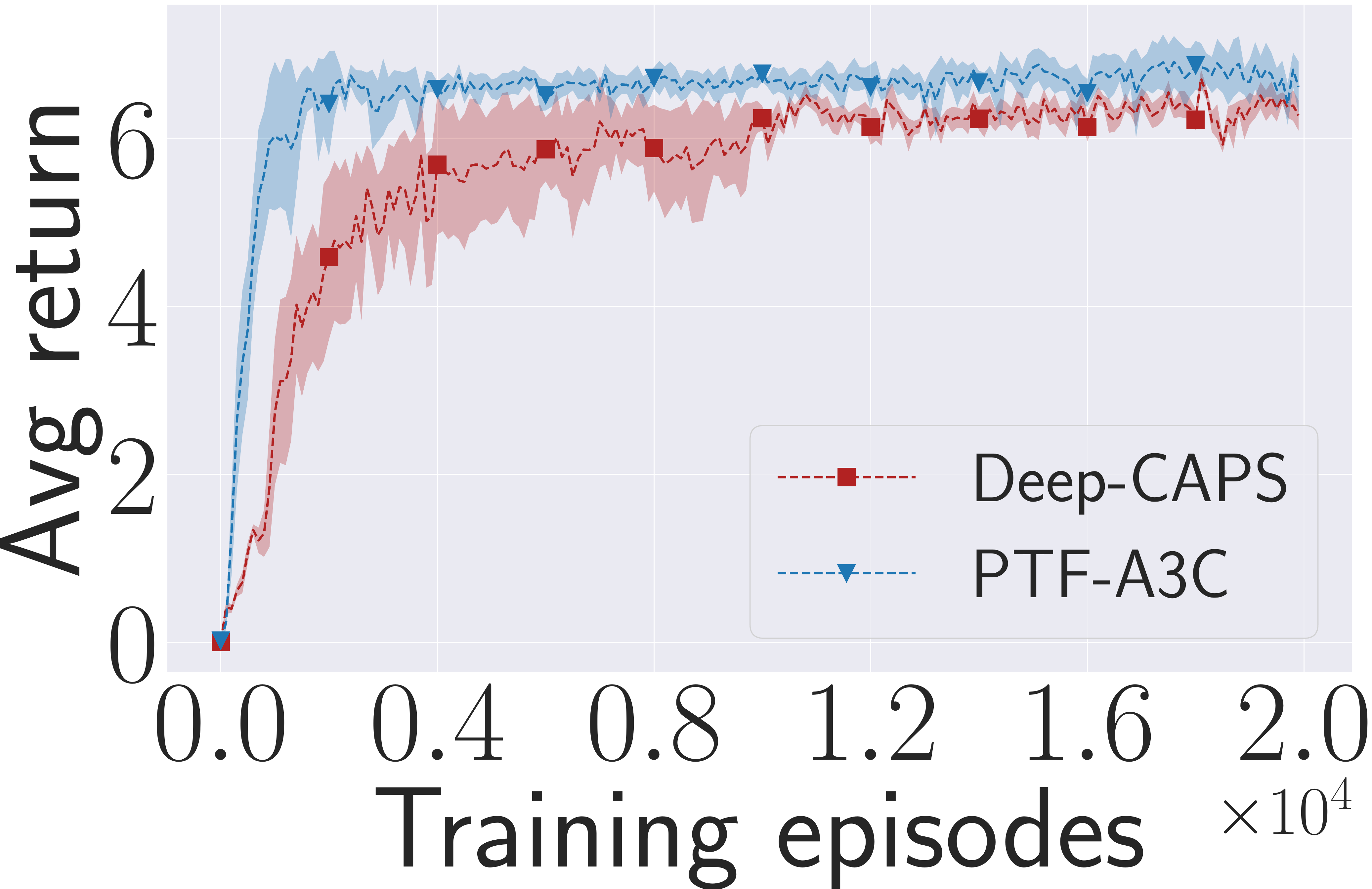}
  }
\caption{Average discounted rewards of various methods when learning $g$ on pinball.} \label{fig-pinball1}
\end{figure*}

\begin{figure*}
\centering
   \subfloat[A3C vs PTF-A3C]{
 \includegraphics[width=.29\linewidth]{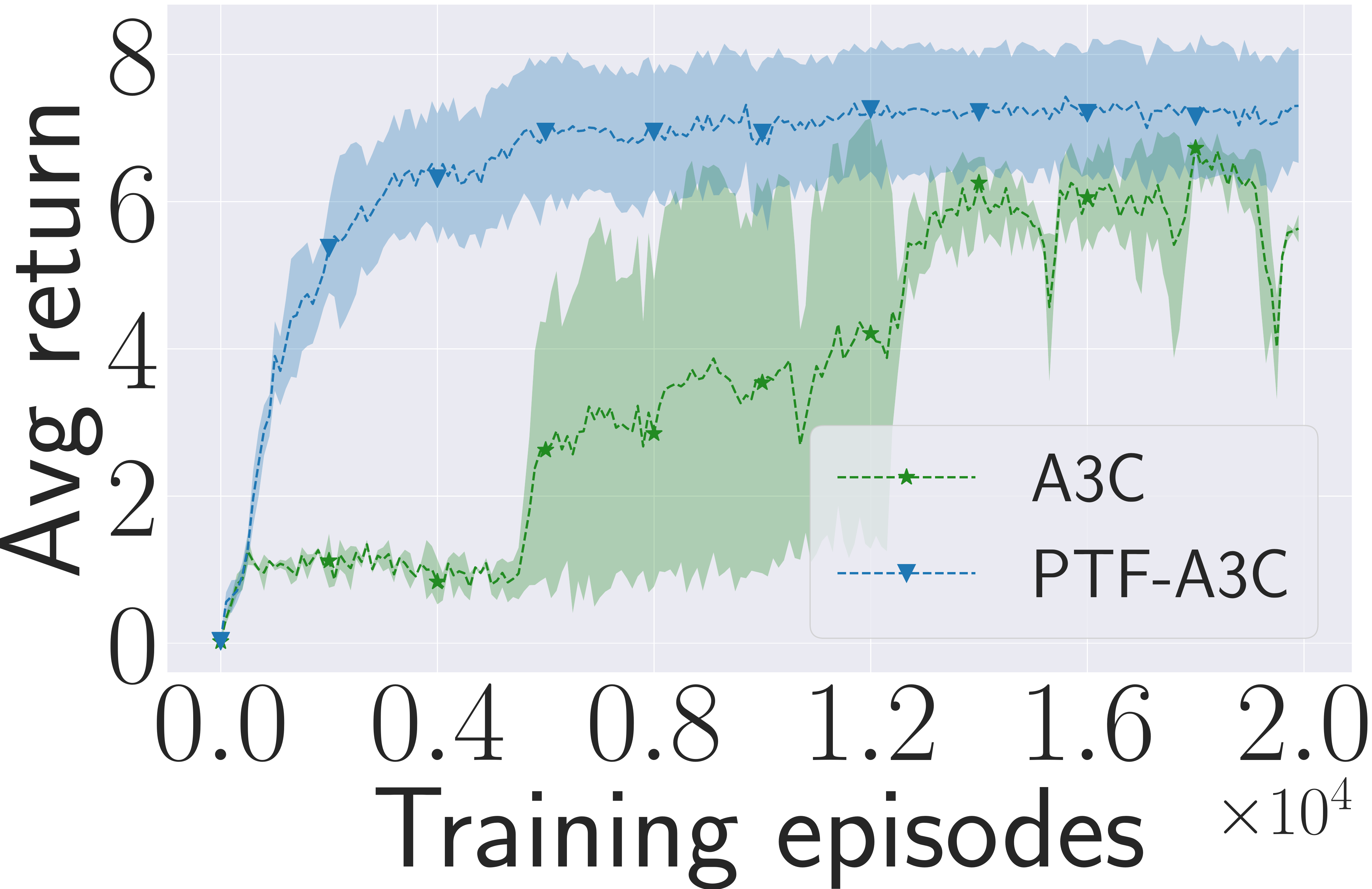}	
  }
   \subfloat[PPO vs PTF-PPO]{ 
	\includegraphics[width=.29\linewidth]{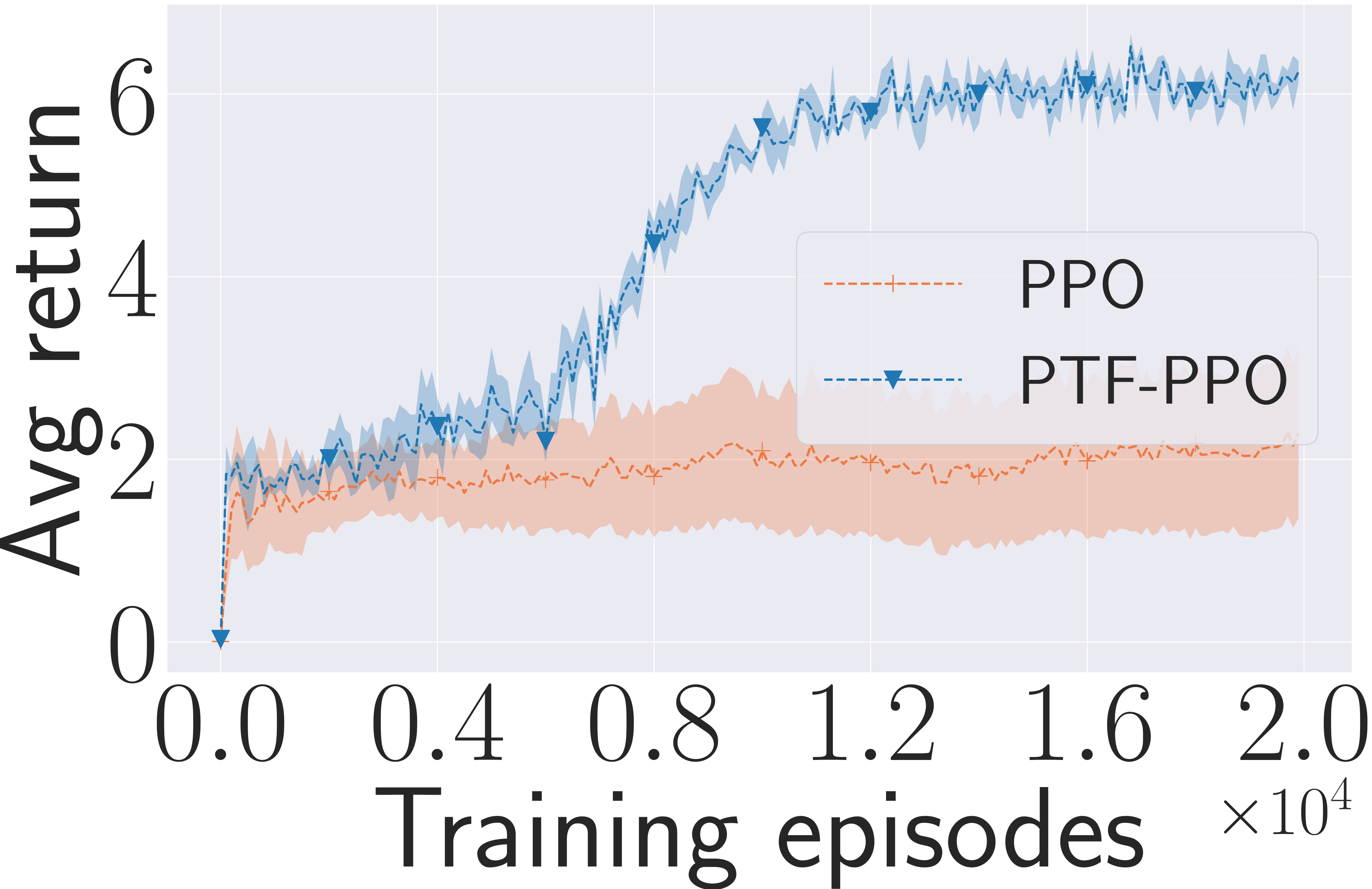}
  } 
   \subfloat[Deep-CAPS vs PTF-A3C]{
  \includegraphics[width=.29\linewidth]{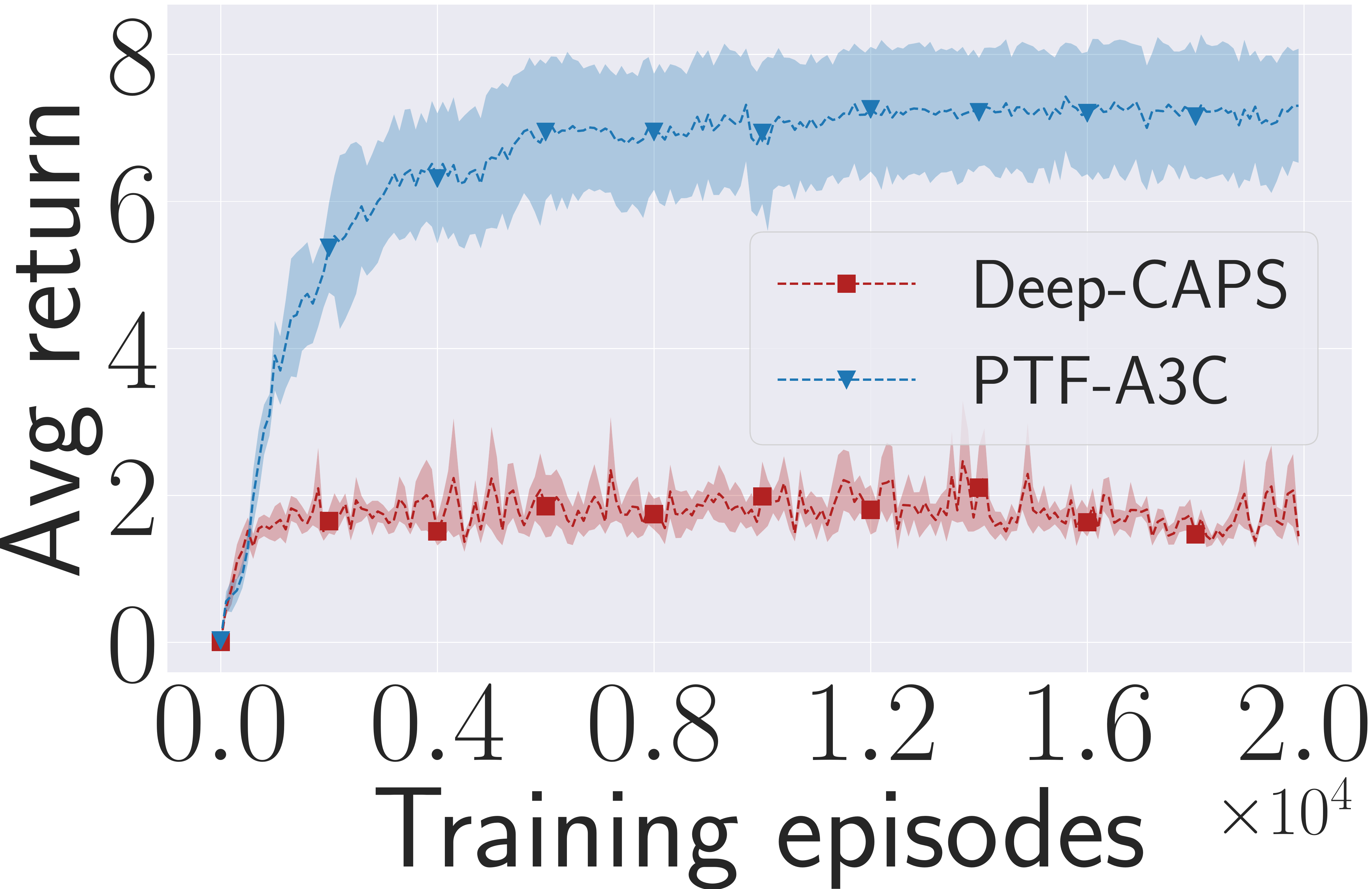}
  }
\caption{Average discounted rewards of various methods when learning $g'$ on pinball.} \label{fig-pinball2}
\end{figure*}

Figure \ref{fig-pinball1} depicts the performance of PTF when learning task $g$ on Pinball, which is similar to source task $G1$ (i.e., the distance between their goal states is very close). We can see that PTF significantly accelerates the learning process of A3C and PPO (Figure \ref{fig-pinball1}(a) and (b)); outperforms deep-CAPS (Figure \ref{fig-pinball1}(c)). The advantage of PTF is similar with that in grid world: PTF efficiently exploits the useful information from source policies to optimize the target policy, thus achieves higher performance than learning from scratch. Deep-CAPS achieves lower average return than PTF but outperforms vanilla A3C and PPO. This indicates it still exploits useful information from source policies when the target task is very similar with one of source tasks. However, it fully depends on source policies for explorations, and a continuous action space is hard to be fully covered even with the manually added primitive policies. Therefore, deep-CAPS achieves lower performance than PTF in such a domain.

We further verify whether PTF works well in the same setting as in the grid world that all source tasks are not quite similar to the target task $g'$ (i.e., the distance between their goal states is very far). From Figure \ref{fig-pinball2} we can see that PTF outperforms other methods even if only some parts of source policies can be exploited. This is because PTF identifies when and which source policy is optimal to exploit and when to terminate it, thus efficiently accelerates the learning process. However, due to the drawbacks described above, deep-CAPS fails when the target task is quite dissimilar with source tasks.

\subsection{Reacher}\label{sec4.1.4}
To further validate the performance of PTF, we provide an alternative scenario, Reacher \cite{dmcontrol}, which is qualitatively different from the above two navigation tasks. Reacher is one of robot control problems in MuJoCo \cite{mujoco}, equipped with a two-link planar to reach a target location. The episode ends with the $+1$ reward when the end effector penetrates the target sphere, or ends when it takes more than $1000$ steps. We design several tasks in Reacher which are different from the location and size of the target sphere. Since deep-CAPS performs poorly in the above continuous domain (pinball), we only compare PTF with vanilla A3C and PPO in the following sections.
\begin{figure}[t]
\centering
\subfloat[A3C vs PTF-A3C]{
\includegraphics[width=0.48\linewidth]{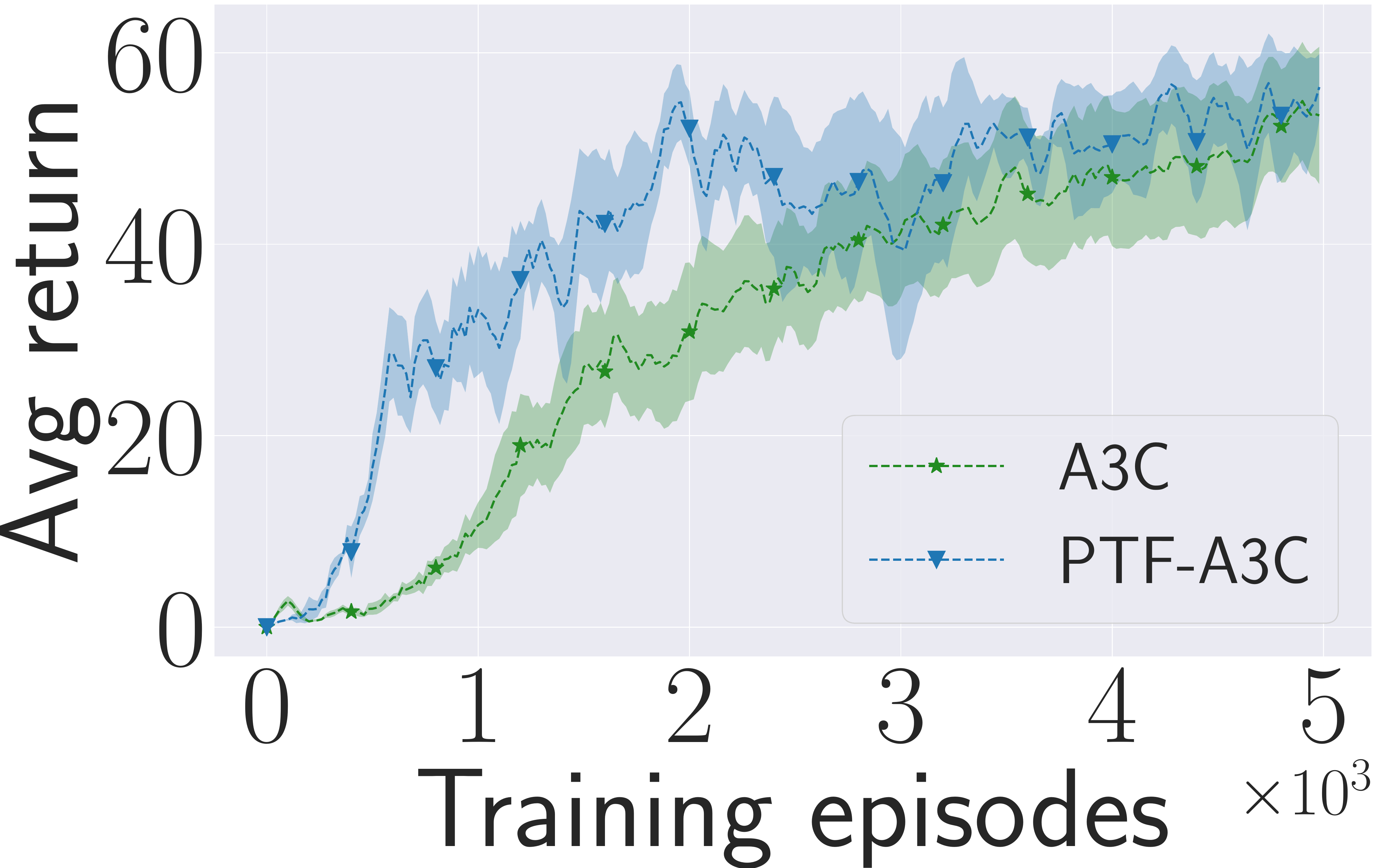}
}
\subfloat[PPO vs PTF-PPO]{
\includegraphics[width=0.48\linewidth]{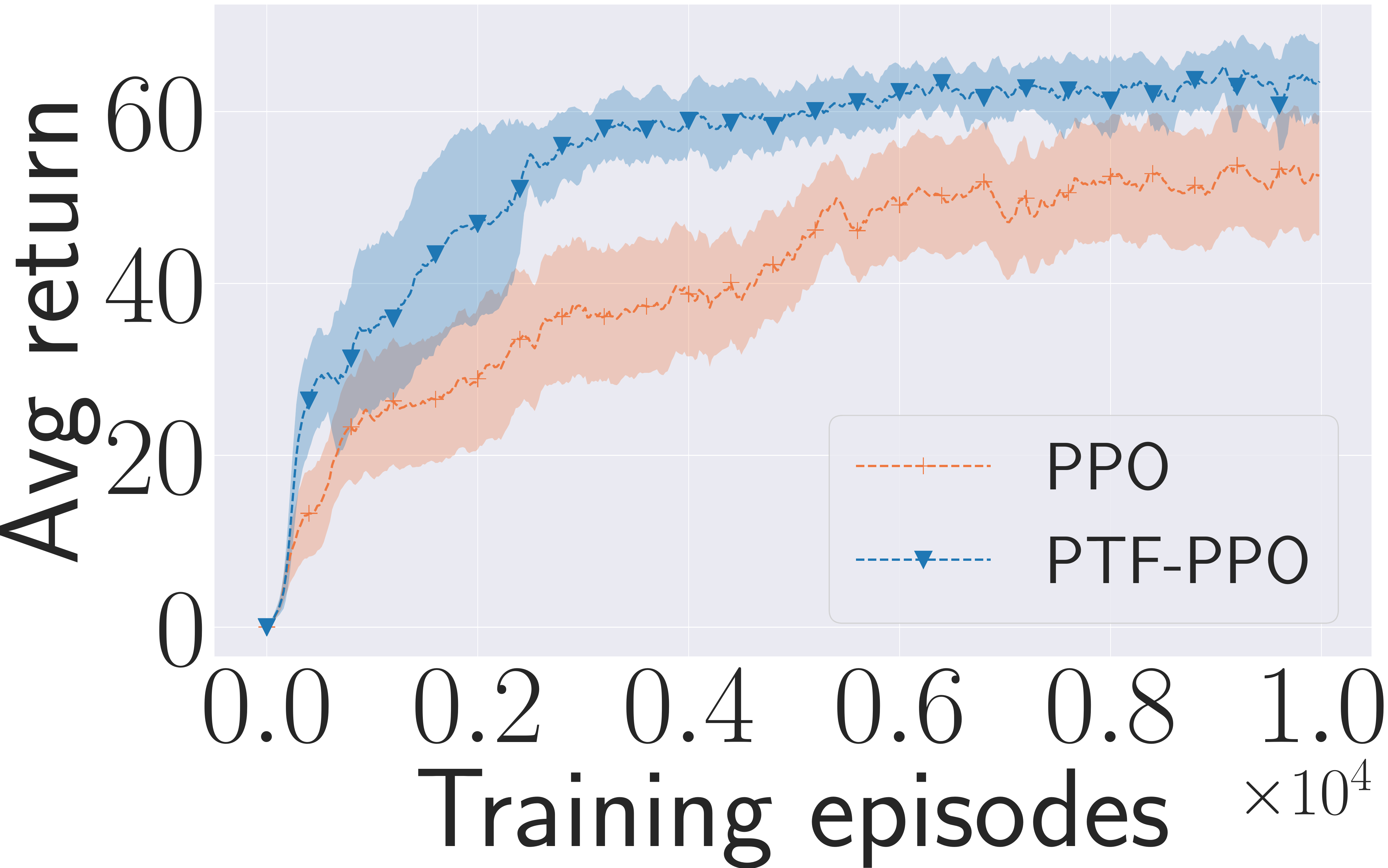}
}
\caption{The performance of PTF on Reacher.} \label{fig-reacher}
\end{figure}
Figure \ref{fig-reacher}(a) shows the performance of PTF-A3C and A3C on Reacher. We can see that PTF-A3C efficiently achieves higher average discounted rewards than A3C. Similar results can be found in PTF-PPO and PPO shown in Figure \ref{fig-reacher}(b). This is because PTF efficiently exploits the useful knowledge in source tasks, thus accelerates the learning process compared with vanilla methods. All results over various environments further show the robustness of PTF.

\subsection{The Influence of $f(\beta_o,t)$}\label{sec4.2}
\begin{figure}[t]
\centering
\includegraphics[width=0.6\linewidth]{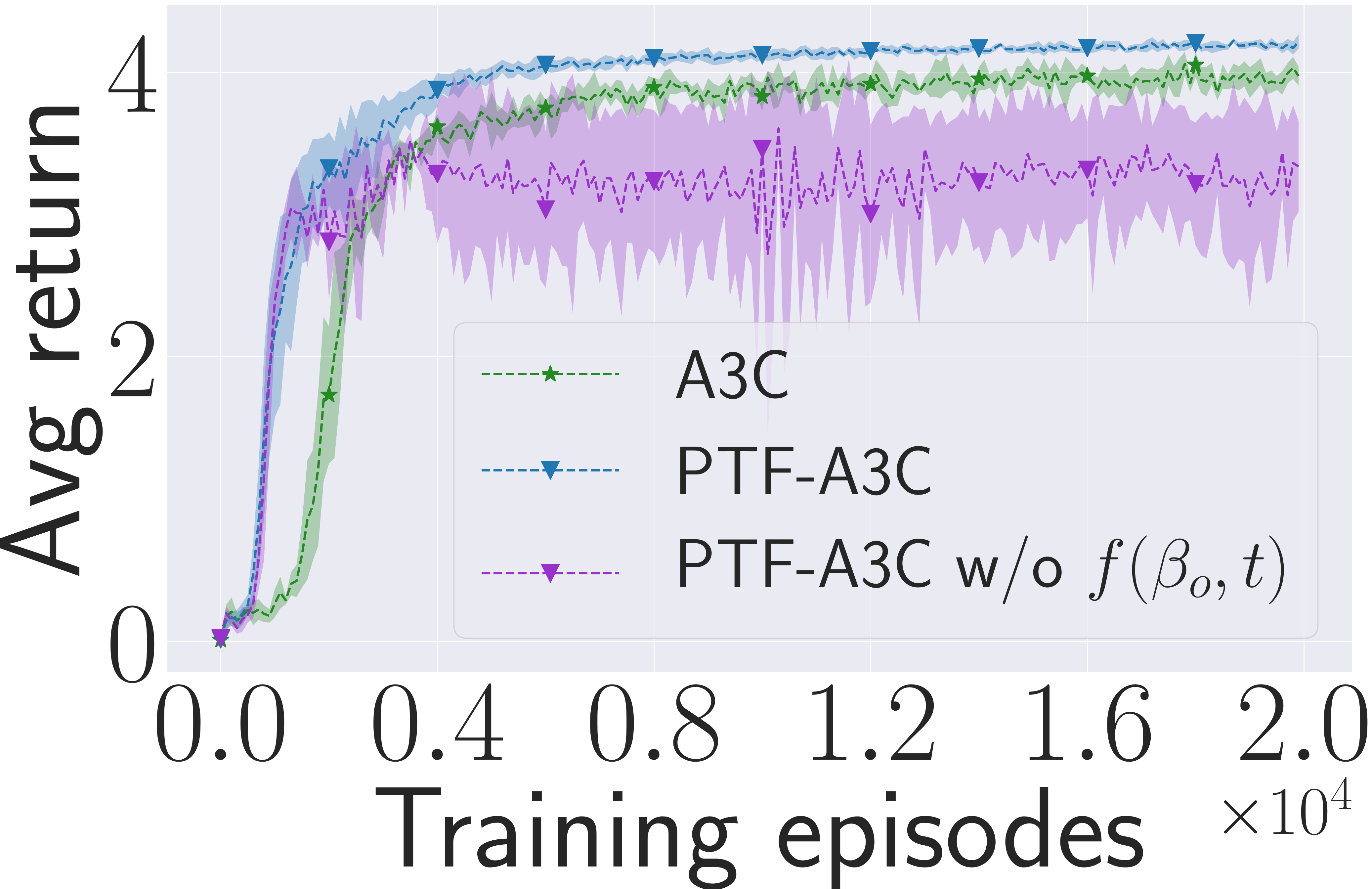}
\caption{The influence of weighting factor $f(\beta_o,t)$.} \label{fig-ablation}
\end{figure}

Next, we provide an ablation study to investigate the influence of the weighting factor $f(\beta_o,t)$ (Equation \ref{eqt}) on the performance of PTF, which is the key factor. Figure \ref{fig-ablation} shows the influence of different parts of the weighting factor on the performance of PTF-A3C. We can see that when the extra loss is added without the weighting factor $f(\beta_o,t)$, although it helps the agent at the beginning of learning compared with A3C learning from scratch, it leads to a sub-optimal policy because of focusing too much on mimicking the source policies. In contrast, introducing the weighting factor $f(\beta_o,t)$ allows us to terminate exploiting source policies in time and thus achieves the best transfer performance.

\subsection{The Performance of Option Learning}\label{sec4.3}
\begin{figure}[t]
\centering
\subfloat[PTF-A3C]
{\includegraphics[width=0.48\linewidth]{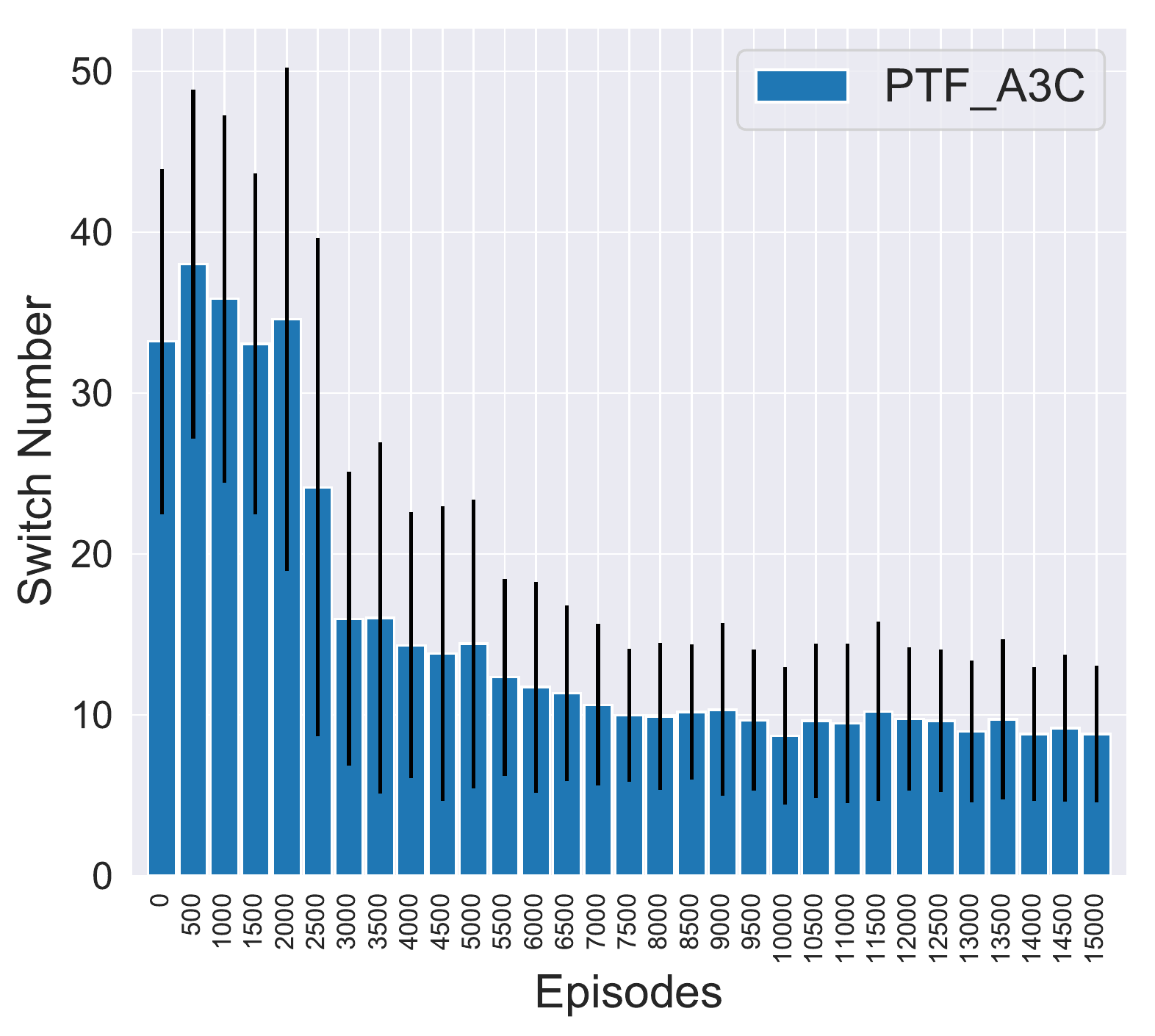}}
\subfloat[PTF-PPO]{\includegraphics[width=0.48\linewidth]{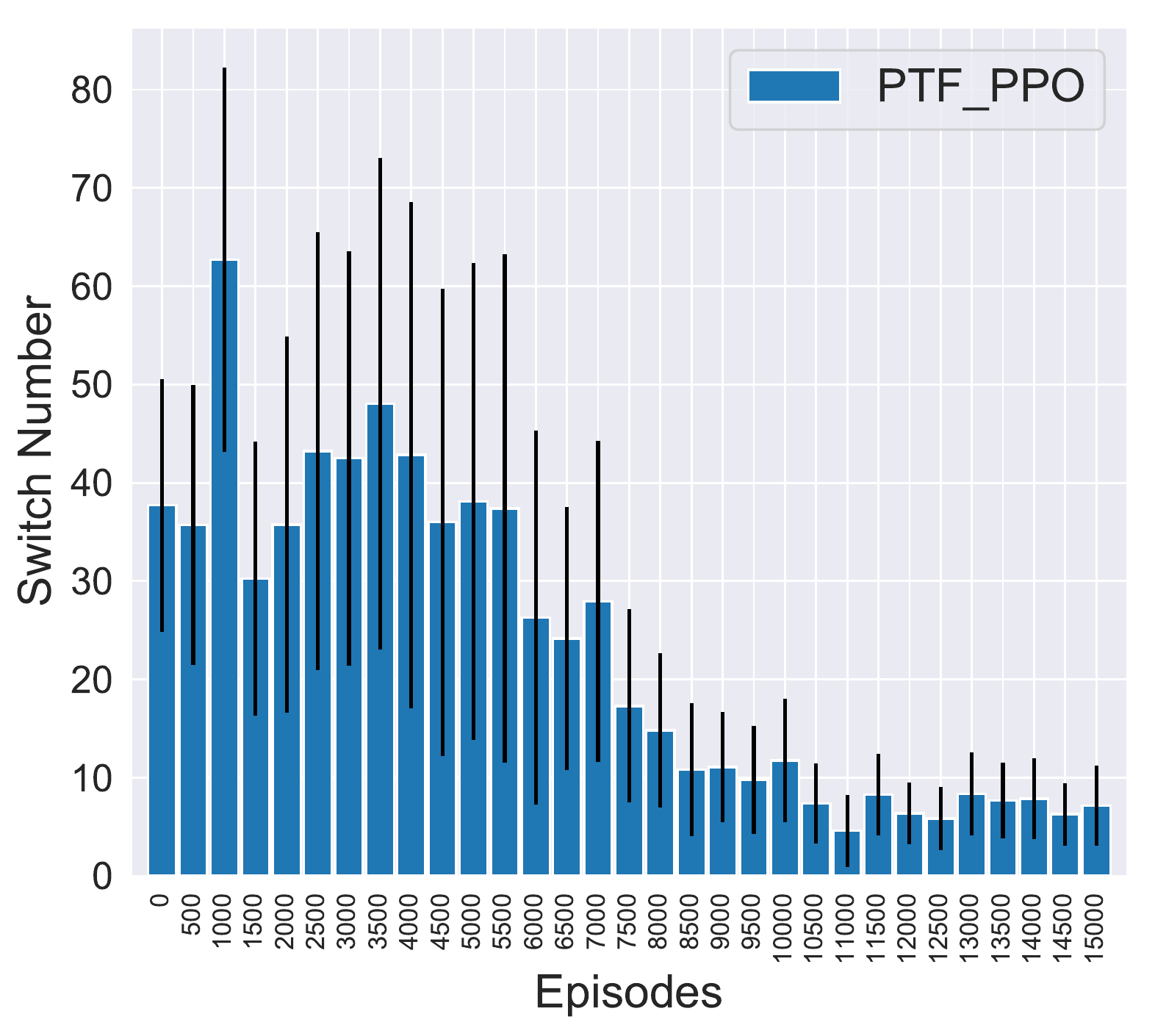}}
\caption{The switch frequency of options.} \label{fig-frequency}
\end{figure}

Finally, we validate whether PTF learns an effective policy over options. Since there may be some concerns about learning termination $\beta_{o}$, that the termination is easy to collapse \cite{bacon2017option,HarutyunyanDBHM19,HarbBKP18}, making it difficult for the policy optimization. In this section, we provide the dynamics of the option switch frequency to investigate the option learning in PTF. From Figure \ref{fig-frequency} (a), (b) we can see that the option switch frequency decreases quickly and stabilizes as the learning goes by. This indicates that both PTF-A3C and PTF-PPO efficiently learn when and which option is useful and provides meaningful guidance for target task learning. 

\section{Conclusion and Future Work}\label{sec5}
In this paper, we propose a Policy Transfer Framework (PTF) which can efficiently select the optimal source policy and exploit the useful information to facilitate the target task learning. PTF also efficiently avoids negative transfer through terminating the exploitation of current source policy and selects another one adaptively. PTF can be easily combined with existing deep policy-based and actor-critic methods. Experimental results show PTF efficiently accelerates the learning process of existing state-of-the-art DRL methods and outperforms previous policy reuse approaches. As a future topic, it is worthwhile investigating how to extend PTF to multiagent settings. Another interesting direction is how to learn abstract knowledge for fast adaptation in new environments.

\section*{Acknowledgments}
The work is supported by the National Natural Science Foundation of China (Grant Nos.: 61702362, U1836214, 61876119), the new Generation of Artificial Intelligence Science and Technology Major Project of Tianjin under grant: 19ZXZNGX00010, and the Natural Science Foundation of Jiangsu under Grant No. BK20181432.


\bibliographystyle{named}  
\bibliography{transfer}  
\section*{Appendix}
\subsection*{Network structure}
The network structure is the same for all methods: the actor network has two fully-connected hidden layers both with 64 hidden units, the output layer is a fully-connected layer that outputs the action probabilities for all actions; the critic network contains two fully-connected hidden layers both with 64 hidden units and a fully-connected output layer with a single output: the state value; the option-value network contains two fully-connected hidden layers both with 32 units; two output layers, one outputs the option-values for all options, and the other outputs the termination probability of the selected option.
\subsubsection*{Grid world} The input consists of the following information: the coordinate of the agent and the environmental information (i.e., each of surrounding eight grids is a wall or not) which is encoded as a one-hot vector. 

\subsubsection*{Pinball} The input contains the position of the ball ($x$ and $y$) and the velocity of the ball in the $x-y$ plane. 

\subsubsection*{Reacher} The input contains the positions of the finger ($x$ and $y$), the relative distance to the target position, and the velocity of in the $x-y$ plane. 
\subsubsection*{Parameter Settings}

\begin{table}[ht]
\caption{CAPS Hyperparameters.}
\centering
\begin{tabular}{c|c}\\
 \hline
\rule{0pt}{12pt}
 Hyperparameter & Value \\
 \hline
 \rule{0pt}{12pt}
 Discount factor($\gamma$) & 0.99 \\
 \rule{0pt}{12pt}
 Optimizer & Adam  \\
 \rule{0pt}{12pt}
 Learning rate& $3e-4$ \\
\rule{0pt}{12pt}
$\epsilon$ decrement & $1e-3$ \\
\rule{0pt}{12pt}
$\epsilon$-start& $1.0$ \\
\rule{0pt}{12pt}
$\epsilon$-end& $0.05$ \\
\rule{0pt}{12pt}
Batch size& $32$ \\
\rule{0pt}{12pt}
Number of episodes \\replacing the target network& $1000$ \\
\hline
\end{tabular}
\end{table}


\begin{table}[ht]
\caption{A3C Hyperparameters.}
\centering
\begin{tabular}{c|c}
 \hline
\rule{0pt}{12pt}
 Hyperparameter & Value \\
 \hline
\rule{0pt}{12pt}
Number of processes & 8 \\
 \rule{0pt}{12pt}
 Discount factor($\gamma$) & 0.99 \\
 \rule{0pt}{12pt}
 Optimizer & Adam  \\
 \rule{0pt}{12pt}
 Learning rate& $3e-4$ \\
\rule{0pt}{12pt}
Entropy term coefficient & $1e-4$ \\
\hline
\end{tabular}
\end{table}

\begin{table}[ht]
\caption{PPO Hyperparameters.}
\centering
\begin{tabular}{c|c}
 \hline
\rule{0pt}{12pt}
 Hyperparameter & Value \\
 \hline
 \rule{0pt}{12pt}
 Discount factor($\gamma$) & 0.99 \\
 \rule{0pt}{12pt}
 Optimizer & Adam  \\
 \rule{0pt}{12pt}
 Learning rate& $3e-4$ \\
 \rule{0pt}{12pt}
 Clip value & $0.2$ \\
\rule{0pt}{12pt}
Entropy term coefficient & 0.005 \\
\hline
\end{tabular}
\end{table}

\begin{table}[ht]
\caption{PTF Hyperparameters.}
\centering
\begin{tabular}{c|c}
 \hline
\rule{0pt}{12pt}
 Hyperparameter & Value \\
 \hline
 \rule{0pt}{12pt}
 Discount factor($\gamma$) & 0.99 \\
 \rule{0pt}{12pt}
 Optimizer & Adam  \\
 \rule{0pt}{12pt}
 Learning rate for the policy network& $3e-4$ \\
 \rule{0pt}{12pt}
 Learning rate for the option network& $1e-3$ \\
\rule{0pt}{12pt}
$f(t)$ & $\frac{1+\tanh (3-0.001t)}{2}$ \\
\rule{0pt}{12pt}
Regularization term  $\xi$ for Equation 5 & $0.001$ \\
\rule{0pt}{12pt}
$\epsilon$ decrement & $1e-3$ \\
\rule{0pt}{12pt}
$\epsilon$-start& $1.0$ \\
\rule{0pt}{12pt}
$\epsilon$-end& $0.05$ \\
\rule{0pt}{12pt}
Batch size& $32$ \\
\rule{0pt}{12pt}
Number of episodes \\replacing the target network& $1000$ \\
\hline
\end{tabular}
\end{table}
\end{document}